\documentclass[runningheads]{llncs}
\usepackage[T1]{fontenc}
\usepackage{graphicx}
\usepackage{booktabs}
\usepackage[misc]{ifsym}
\newcommand{\corr}{(\Letter)}

\usepackage{algorithmic}
\usepackage{textcomp}
\usepackage{xcolor}
\usepackage{amsmath}
\usepackage{makecell, multirow, array}
\usepackage{amssymb,amsfonts}
\usepackage{subcaption}

\usepackage{pifont}
\usepackage{tcolorbox}
\newcommand{\cmark}{\ding{51}}
\newcommand{\xmark}{\ding{55}}

\begin{document}

\title{VLC Fusion: Vision-Language Conditioned \\Sensor Fusion for Robust Object Detection}

\titlerunning{Vision-Language Conditioned Sensor Fusion for Robust Object Detection}



\author{ 
Aditya Taparia\inst{1}~\corr \and 
Noel Ngu\inst{1} \and 
Mario Leiva\inst{2} \and 
Joshua Shay Kricheli\inst{5} \and 
John Corcoran\inst{3} \and 
Nathaniel D. Bastian\inst{4} \and 
Gerardo Simari\inst{2} \and 
Paulo Shakarian\inst{5} \and 
Ransalu Senanayake\inst{1} 
} 
\authorrunning{A. Taparia et al.} 

\institute{Arizona State University, Tempe, AZ, USA\\
\email{\{ataparia,nngu2,ransalu\}@asu.edu}
\and Dept.\ of Computer Science and Engineering, Universidad Nacional del Sur, Institute for Computer Science and Engineering, Bah\'ia Blanca, Argentina\\
\email{\{mleiva3,gsimari\}@asu.edu}
\and U.S.\ Department of Defense, Arlington, VA, USA\\
\email{john.corcoran.ctr@darpa.mil}
\and United States Military Academy, West Point, NY, USA\\
\email{nathaniel.bastian@westpoint.edu}
\and Syracuse University, Syracuse, NY, USA\\
\email{\{jkrichel,pashakar\}@syr.edu}}



\toctitle{VLC Fusion: Vision-Language Conditioned Sensor Fusion for Robust Object Detection}
\tocauthor{Aditya Taparia, Noel Ngu, Mario Leiva, Joshua Shay Kricheli, John Corcoran, Nathaniel D. Bastian, Gerardo Simari, Paulo Shakarian, Ransalu Senanayake}

\maketitle              

\begin{abstract}
Although fusing multiple sensor modalities can enhance object detection performance, existing fusion approaches often overlook subtle variations in environmental conditions and sensor inputs. As a result, they struggle to adaptively weight each modality under such variations. To address this challenge, we introduce Vision-Language Conditioned Fusion (VLC Fusion), a novel fusion framework that leverages a Vision-Language Model (VLM) to condition the fusion process on nuanced environmental cues. By capturing high-level environmental context such as darkness, rain, and camera blurring, the VLM guides the model to dynamically adjust modality weights based on the current scene, ensuring robustness against environmental shifts. We evaluate VLC Fusion on real-world autonomous driving and military target detection datasets that include image, LIDAR, and mid-wave infrared modalities. Our experiments show that VLC Fusion consistently outperforms conventional fusion baselines, achieving improved detection accuracy in both seen and unseen scenarios. \textbf{Supplementary Material and Code:} \url{https://github.com/aditya-taparia/VLCFusion}

\keywords{Sensor Fusion  \and Vision-Language Models \and Object Detection.}
\end{abstract}

\section{Introduction}

\begin{figure}[t]
    \centering
    \includegraphics[width=\linewidth]{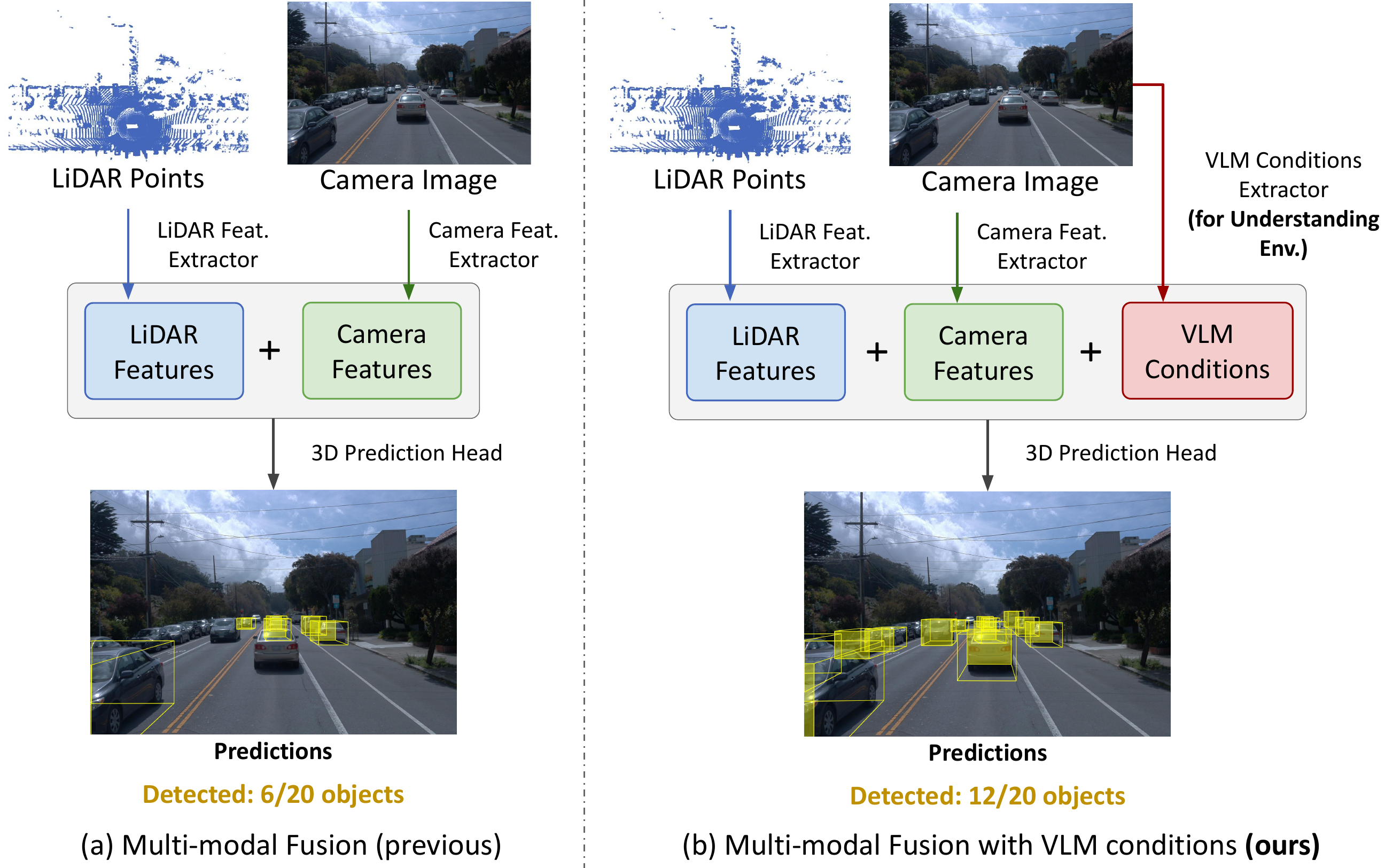}
    \caption{Overview of \textbf{VLC Fusion}. Compared to (a) standard fusion for object detection, (b) our method modulates modality-specific features with environment-specific meta-information, called \emph{conditions}, improving the resilience of object detection to diverse natural environmental variations.}
    \label{fig:vlm_fusion}
\end{figure}

\begin{figure*}
    \centering
    \includegraphics[width=\textwidth]{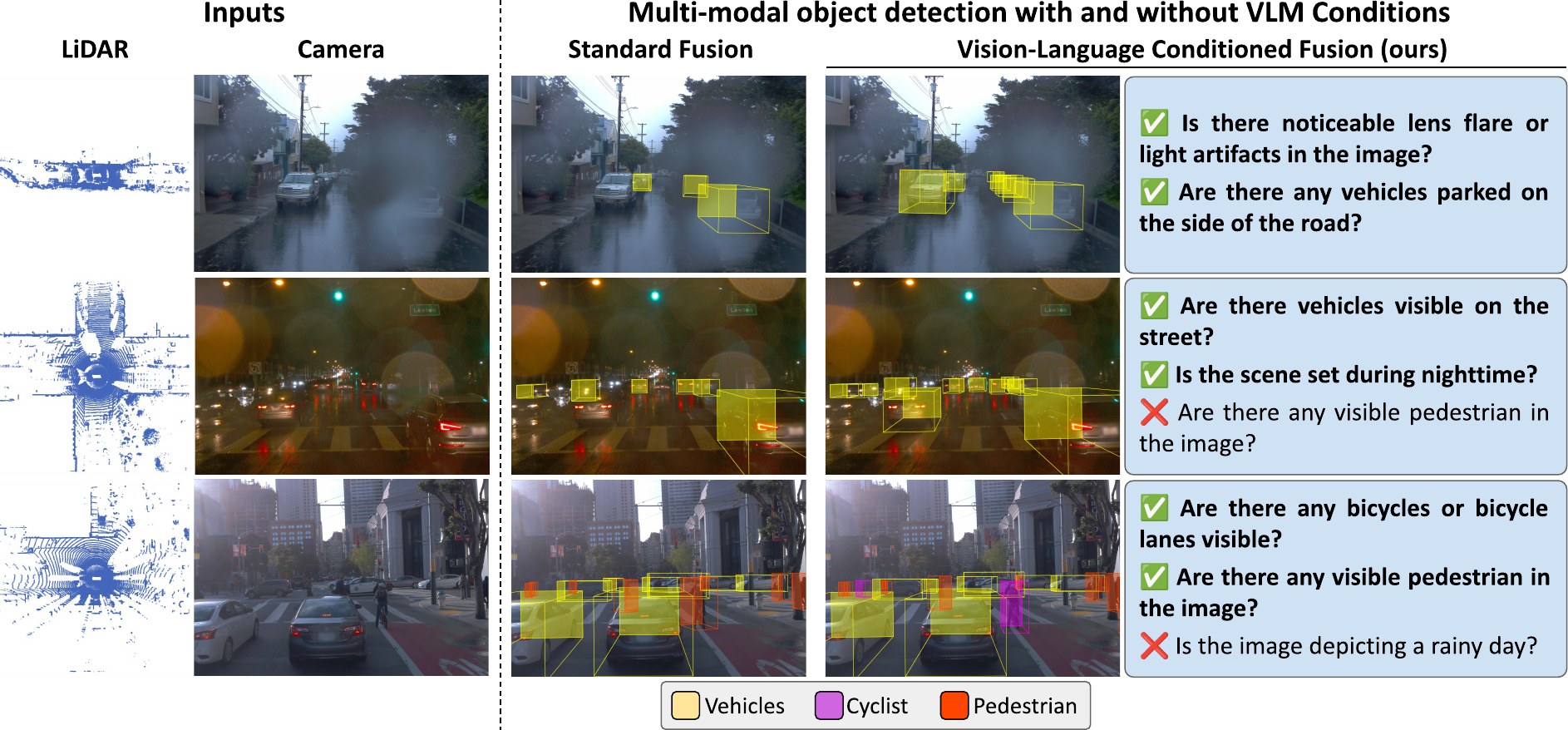}
    \caption{Comparison of sample predictions from multi-modal fusion and VLC Fusion: from left to right, raw LiDAR point clouds and camera views, predictions from a multi‐modal fusion baseline, and predictions from our VLC Fusion conditioned on VLM‐extracted environmental cues (\cmark/\xmark prompts shown at right). Conditioning on high‐level context improves detection performance in recovering occluded cars (Example 1), detecting more vehicles under nighttime glare (Example 2), and correctly identifying the cyclist (magenta) and pedestrians (orange) (Example 3) where the baseline fails.}
    \label{fig:intro_image}
\end{figure*}

Reliable object detection is critical for many real-world autonomous systems such as autonomous vehicles and surveillance platforms. Since different sensor modalities offer distinct advantages, multi-modal fusion techniques aim to integrate object detectors trained on these different modalities. For example, since RGB images provide high-resolution detail while LIDAR offers depth perception despite its sparse point cloud, sensor fusion can provide a high-resolution image with some depth information.

A key limitation of current fusion methods is that they overlook how the performance of each modality varies with external environmental conditions. Since object detection models are optimized individually for specific sensor modalities, each excels under certain environmental conditions but exhibits vulnerabilities under others. For instance, state-of-the-art RGB-based object detectors such as Detection Transformer (DETR)~\cite{carion2020end} performs well in clear, well-lit conditions but degrade considerably in low-light or adverse weather scenarios such as fog~\cite{bijelic2020seeing,pathiraja}. Conversely, LiDAR-based object detectors such as PointPillars~\cite{lang2019pointpillars} and SECOND~\cite{yan2018second} provide robust performance under varied lighting levels but can deteriorate in weather conditions such as rain due to light scattering and other sensor-specific limitations~\cite{delecki2022we}. The problem of environment dependence becomes even more pronounced when the system is deployed in unseen environments.

To address this challenge, we propose Vision-Language Conditioned Fusion (VLC Fusion), a novel approach to incorporate environmental meta-information obtained through Vision-Language models (VLMs) into the fusion process. Since current state-of-the-art VLMs have demonstrated impressive scene‑understanding capabilities~\cite{li2023blip,liu2023visual}, we use them offline to reliably extract detailed environmental cues across a wide range of real-world tasks. We then proposed an architecture to incorporate the VLM conditions into the fusion network. At test-time, in addition to raw sensor inputs, the VLM provides an analysis of the scene, making our method robust to both seen (in‑distribution) and unseen (out‑of‑distribution) scenarios. Fig.~\ref{fig:vlm_fusion} illustrates the difference between (a) standard multi‑modal fusion and (b) our VLC Fusion. Furthermore, Fig.~\ref{fig:intro_image} presents qualitative examples where this conditioning on VLM cues leads to better detection. The primary contributions of the paper are:

\begin{enumerate}
    \item We propose a novel fusion approach, called VLC Fusion, that automatically weighs feature fusion on environment‑specific meta‑information.
    \item We introduce an automated framework for offline extraction and integration of relevant environmental cues from raw datasets.
    \item We empirically demonstrate the usefulness of environmental conditions in multi-modal sensor fusion on two real-world object detection tasks, autonomous driving and military target detection. We also demonstrate how lightweight, fast, small-scale VLMs can be used realistically during the online object detection phase.
\end{enumerate}
\section{Related Work}

\textbf{Multi‐Modal Sensor Fusion for Object Detection.}
LiDAR-camera fusion methods demonstrated clear benefits over single‐modality approaches by combining precise geometric measurements with dense visual context. Previous works such as 
AVOD \cite{ku2018joint} project LiDAR point clouds into the image plane to jointly learn features. While PointFusion \cite{xu2018pointfusion} fuses raw point embeddings with image features via a learned weighting scheme. Fusion SSD \cite{bijelic2020seeing} concatenates feature maps from both modalities and applies convolutional layers for joint detection, and Learnable Align \cite{li2022deepfusion} uses a cross‐attention block to align and integrate modality‐specific features. More recent works such as TransFusion \cite{bai2022transfusion} and PillarNeXt \cite{li2023pillarnext} use cross‐modal attention to further improve alignment at multiple scales. Although these methods achieve good performance, they generally \textit{apply static fusion rules} that do not adjust to changing environmental conditions \cite{bijelic2020seeing}. This lack of adaptability leads to degraded performance when encountering conditions not well‐represented in the training data, such as sudden changes in illumination or unusual weather patterns \cite{delecki2022we,pathiraja}. We propose the use of environmental‐cues to dynamically weigh the importance of each modality.

\textbf{Condition‐Aware Fusion Approaches.}
To handle diverse lighting and weather scenarios, subsequent work recognized the need for adaptability, introducing condition‐aware mechanisms that adapt fusion weights based on environment estimates.
DSFuse \cite{li2020ds} uses uncertainty estimates to downweight noisy modalities. More recently,
RGB-X~\cite{deevi2024rgb} proposes the use of scene agnostic switch to switch between detection head based on particular scenario. CAFuser \cite{broedermann2023cafuser} proposes a learned condition token trained with a CLIP style loss to embed discrete scene types (e.g., “rainy”, “foggy”) and guide multi-modal feature fusion. Despite these advances, most methods \textit{rely on a fixed taxonomy of conditions and require annotated examples} for each. This reliance on predefined categories restricts their ability to handle ambiguous or continuously varying conditions (e.g., light fog transitioning to heavy fog) and requires potentially expensive data annotation efforts for every new condition, limiting their flexibility when encountering novel or mixed scenarios \cite{broedermann2023cafuser}. On the contrary, we create application-specific conditions and also provide a way to automatically identify these relevant conditions.

\textbf{Vision–Language Models for Context‐Awareness.}
Building on condition aware mechanisms, recent work explores the use of large pre-trained vision language models (VLMs) for extracting semantic context. Models like CLIP~\cite{radford2021learning} support matching image regions to arbitrary text, while MDETR~\cite{kamath2021mdetr} extends this to end-to-end phrase grounding. More integrated approaches, such as PaLM-E~\cite{driess2023palm} and RoboFlamingo~\cite{li2023vision}, combine vision, language, and robot state for downstream reasoning. These models enable systems to understand not just what objects are present, but also how they relate to tasks or environmental cues. However, leveraging VLMs to infer environmental cues (e.g., “bright urban afternoon” vs. “dusty desert dusk”) rather than relying on fixed, discrete categories and using these insights to guide real-time sensor fusion remains an open challenge. Our work addresses this gap by using a pretrained VLM to extract environmental conditions that adaptively modulate sensor weighting, enabling context-aware fusion without requiring explicit labels.

\section{Methodology}

Our methodology comprises two major components: 1) identification of application specific environmental conditions and querying VLM to obtain the corresponding responses, and 2) integrating these conditional information into the sensor fusion architecture. We first describe how application-specific conditions are identified and queried from VLMs, and then detail how the resulting conditional information is integrated into the fusion network.

\subsection{Offline Condition Extraction and Generation}
\label{sec:condition_extraction}
Extracting meaningful environmental conditions is crucial for guiding sensor fusion in our method. To this end, we explored two ways by which one can extract (or define) conditions from a dataset.

\textbf{Human-Defined Conditions.} Leveraging prior domain knowledge and metadata available from dataset, experts can manually define relevant conditions based on the application. While straightforward, this approach can be subjective and may not generalize effectively across diverse datasets and environments.

\begin{figure}[ht]
    \centering
    \includegraphics[width=\linewidth]{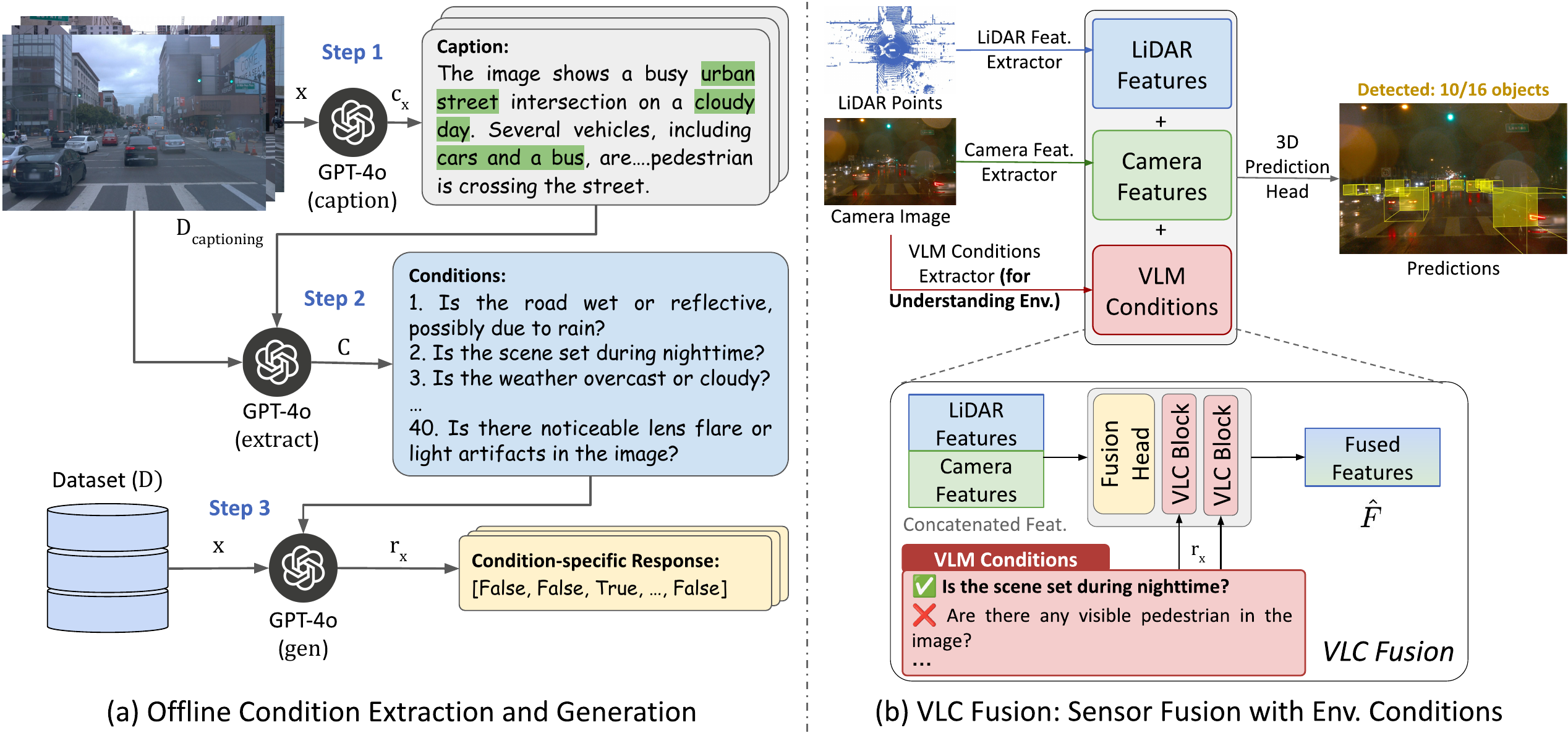}
    \caption{Overview of (a) the three-step automated pipeline for extracting environmental conditions: captioning a subset of images, extracting a set of conditions, and querying the VLM to generate a binary condition-response vector $r_x$ per image. (b) Sensor fusion with environmental conditions, where stacked VLC blocks modulate the fused features using $r_x$ before the detection head.}
    \label{fig:vlm_condition_extractor}
\end{figure}

\textbf{Automated Condition Extraction.} To overcome limitations associated with manual definitions, we introduce an automated framework for offline extraction of rich environmental information and contextual cues from dataset via VLM. For this purpose, as shown in Fig.~\ref{fig:vlm_condition_extractor}a, we introduce a three step process:

\paragraph{Step 1 (Captioning).}
Let the training dataset be \(D\) with \(N\) images, and let
\(D_{\mathrm{captioning}} \subset D\) denote a randomly selected subset of
\(M\) images. Using a captioning function
\(\mathrm{caption}(\cdot; p_{\mathrm{caption}})\), we generate a descriptive
caption \(c_x\) for each image \(x \in D_{\mathrm{captioning}}\):
\[
c_x = \mathrm{caption}(x; p_{\mathrm{caption}}), \qquad \forall x \in D_{\mathrm{captioning}} .
\]
Here, \(p_{\mathrm{caption}}\) is the text prompt
``Describe the input scene,'' together with the system prompt provided in
Supplementary. This gives us \(M\) image-caption pairs
\((x, c_x)\).

\paragraph{Step 2 (Extraction).}
After generating captions, we apply \(\mathrm{extract}(\cdot)\) to the
\(M\) image-caption pairs to derive a set of environmental conditions
\(C\):
\[
C \leftarrow \mathrm{extract}\big(\{(x_m, c_{x_m})\}_{m=1}^M;\,
p_{\mathrm{extraction}}\big).
\]
Here, \(p_{\mathrm{extraction}}\) is the prompt ``Provide conditions based on the following image--caption pairs,'' used together with the system prompt described in Supplementary. This step converts free-form captions into a structured set of application-specific environmental conditions. For example, from the caption \textit{``busy urban intersection on a cloudy day,''} the extracted conditions may include \textit{``presence of vehicles,''} \textit{``cloudy weather,''} and \textit{``busy pedestrian activity.''} In this way, the extraction stage captures both high-level scene semantics and fine-grained contextual cues. Before using these conditions for training and evaluation, we remove duplicate entries.

\paragraph{Step 3 (Generation).}
Once the environmental conditions have been identified, we query a pretrained VLM to determine whether each condition is present in each image from the full training dataset \(D\). Concretely, for every image \(x \in D\) and condition \(c \in C\), we generate a binary response:
\[
r_{x,c} = \mathrm{gen}(x,c), \qquad \forall x \in D,\; c \in C,
\]
where \(r_{x,c} \in \{\mathrm{True}, \mathrm{False}\}\) indicates whether condition \(c\) is present in image \(x\). In our implementation, GPT-4o is used for this condition-response generation step. For a given image
\(x\), this produces a condition-response vector:
\[
r_x = [r_{x,c_1}, r_{x,c_2}, \dots, r_{x,c_{|C|}}].
\]
This vector serves as the environmental condition representation for image \(x\) and is used as input to the fusion model during training and
evaluation.

\subsection{Sensor Fusion with Environmental Conditions}
\label{sec:fusion}

We now describe how the queried environmental condition vector is integrated into the fusion network. Let \(F_1 \in \mathbb{R}^{B \times C_1 \times H \times W}\) and \(F_2 \in \mathbb{R}^{B \times C_2 \times H \times W}\) denote feature maps from two sensing modalities, such as visible and infrared imagery or RGB and LiDAR features. Let \(r_x \in \{0,1\}^{|C|}\) denote the environmental condition vector obtained for input \(x\) using the procedure in Section~\ref{sec:condition_extraction}. Our goal is to learn an environment-conditioned fusion function:
\[
\mathcal{F}(F_1,F_2,r_x) \mapsto \hat{F},
\]
where \(\hat{F}\) is the fused representation passed to the downstream detection head.

\textbf{Fusion Head.}
As shown in Fig.~\ref{fig:vlm_condition_extractor}b, we first concatenate the modality-specific feature maps:
\[
U = [F_1;F_2] \in \mathbb{R}^{B \times (C_1+C_2) \times H \times W}.
\]
This concatenated representation \(U\) is passed through a multi-scale fusion head composed of three parallel convolutional branches with kernel sizes \(1\times1\), \(3\times3\), and \(5\times5\). Their outputs are combined using input-adaptive weights to produce an initial fused feature map \(F \in \mathbb{R}^{B \times C_{\mathrm{out}} \times H \times W}\). 

\begin{figure}[t]
    \centering
    \begin{minipage}[t]{0.6\linewidth}
        \vspace{0pt}
        \centering
        \captionof{table}{Dataset splits for the Waymo Open and ATR datasets across seen and unseen scenarios. For Waymo, the seen split covers day and night driving while the unseen split covers dawn and dusk. For ATR, the seen split covers object distances of 1000--5000 m while the unseen split covers intermediate distances (1500 m, 2500 m, 3500 m, and 4500 m). The unseen scenarios are held out exclusively for evaluation.}
        \label{tab:dataset_splits_combined}
        \renewcommand{\arraystretch}{1.2}
        \resizebox{\linewidth}{!}{%
        \begin{tabular}{l|c|c|c|c}
            \toprule
            Dataset & Variation & Train & Validation & Test \\
            \midrule
            \multirow{2}{*}{\makecell[l]{Waymo Open dataset\\(RGB + LiDAR)}}
                & Seen & 73,112 & 9,139 & 9,139 \\
                & Unseen & -- & -- & 7,052 \\
            \midrule
            \multirow{2}{*}{\makecell[l]{ATR dataset\\(Visible + MWIR)}}
                & Seen & 45,207 & 15,075 & 15,088 \\
                & Unseen & -- & -- & 11,952 \\
            \bottomrule
        \end{tabular}}
    \end{minipage}
    \hfill
    \begin{minipage}[t]{0.35\linewidth}
        \vspace{0pt}
        \centering
        \includegraphics[width=0.85\linewidth]{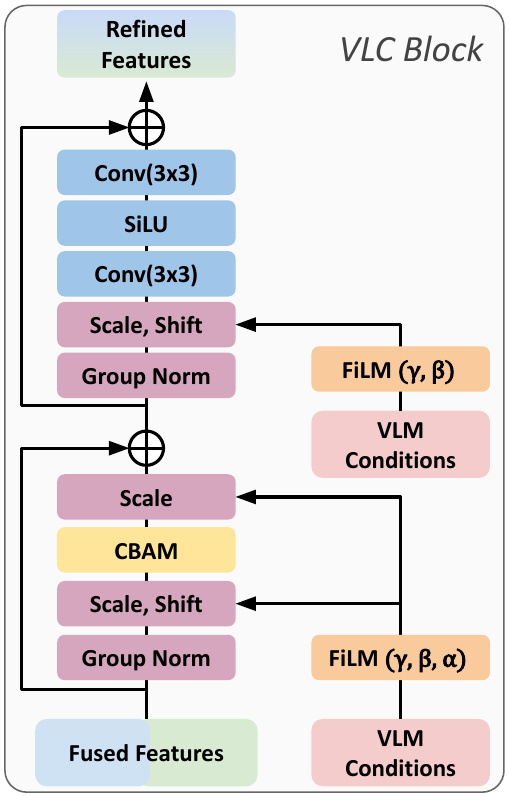}
        \captionof{figure}{Architecture of the VLC Block.}
        \label{fig:vlc_block}
    \end{minipage}
\end{figure}

\textbf{VLC Block.}
As shown in Fig.~\ref{fig:vlm_condition_extractor}b, the fused feature \(F\) is refined using our VLC block (see Fig.~\ref{fig:vlc_block}), which combines normalization, FiLM-based conditioning~\cite{perez2018film}, CBAM attention~\cite{woo2018cbam}, and residual convolutional refinement. The purpose of the VLC block is to adapt the fused representation according to the queried environmental conditions.

Given an input feature map \(Z \in \mathbb{R}^{B \times C \times H \times W}\), the VLC block first maps the environmental condition vector \(r_x\) to channel-wise FiLM parameters:
\[
(\gamma,\beta,\alpha)=\phi(r_x),
\qquad
\gamma,\beta,\alpha \in \mathbb{R}^{B \times C \times 1 \times 1}.
\]
It then normalizes the fused representation using Group Norm (GN), applies condition-dependent affine modulation, and 
refines the result using Convolutional Block Attention Module (CBAM):
\[
\begin{aligned}
\bar{Z} &= (1+\gamma)\odot \mathrm{GN}(Z) + \beta \\
Z' &= Z + \alpha \odot \mathrm{CBAM}(\bar{Z}).
\end{aligned}
\]
Here, \(\gamma\) and \(\beta\) perform condition-dependent channel-wise scaling and shifting, while \(\alpha\) controls the contribution of the attention-refined feature to the residual update. This design allows the model to condition the fused features on the environmental vector and focus on relevant features.

To further improve representation quality, the VLC block includes a second residual refinement stage using a lightweight convolutional feed-forward network. Concretely, \(Z'\) is normalized and modulated by another FiLM transformation, and then passed through a feed-forward convolutional block. Unlike the first stage, this refinement step does not use a FiLM-based control gate in the end:
\[
\begin{aligned}
\bar{Z}' &= (1+\gamma_f)\odot \mathrm{GN}(Z') + \beta_f, \\
\mathrm{VLC}(Z,r_x) &= Z' + \mathrm{FFN}(\bar{Z}').
\end{aligned}
\]

\textbf{VLC Fusion Architecture.}
In the complete architecture, the output of the fusion head, $F$ is passed through two VLC blocks sequentially:
\[
\hat{F} = \mathrm{VLC}_2(\mathrm{VLC}_1(F, r_x), r_x).
\]
The first VLC block performs condition aware adjustment of the initial multimodal representation, while the second block refines the fused feature before it is fed to the detector. The final fused representation \(\hat{F}\) is then passed to the task-specific detection head. We inject environmental conditions after initial multimodal aggregation so that the unimodal backbones preserve stable modality-specific representations, while the joint representation can be adaptively reweighted according to the sensing environment.

\section{Experiments}

In this section, we empirically evaluate our proposed VLC Fusion methodology using two real-world datasets: the Waymo Open dataset~\cite{sun2020scalability}, where we fuse RGB and LiDAR modalities for 3D object detection, and the Automated Target Recognition (ATR) dataset~\cite{atr2010}, where we fuse visible and infrared (IR) imagery for 2D object detection. We evaluate the effectiveness of VLC Fusion under both seen (training distribution) and unseen (out-of-distribution) scenarios. As shown in Supplementary, 85\% and 76\% of samples in the Waymo and ATR datasets, respectively, contain at least one active environmental condition.
Below, we first detail the creation of seen and unseen datasets, followed by metrics, implementation details, baseline methods, results, and ablations.

\subsection{Datasets}
\textbf{Waymo Open Dataset:} For our experiments, we use the San Francisco portion of the Waymo Open dataset, which provides synchronized LiDAR and RGB imagery captured at 10 Hz in a busy urban environment with diverse weather (e.g. rainy, sunny) and lighting conditions (e.g. day time, night time, dawn/dusk time). Each sequence contains approximately 200 frames across 20 seconds.

We define two scenarios: a seen scenario, which includes data collected under day and night time, and an unseen scenario, comprising data from dawn and dusk. The seen scenario was used for both training and testing, while the unseen scenario was reserved strictly for testing. Prior to training, frames were shuffled to ensure diversity and robustness. The resulting train-val-test splits for both scenarios are summarized in Table~\ref{tab:dataset_splits_combined}.

\textbf{ATR Dataset:} The ATR dataset contains visible and mid-wave infrared (MWIR) imagery aimed at target recognition applications, and comprehensive metadata detailing object distances, viewing angles, wind speed, and other relevant attributes.
We first synchronized the frames from the two modalities using timestamps and object metadata to ensure proper alignment. Following synchronization, we partitioned the dataset into seen and unseen scenarios based on object distance. The seen set includes distances of 1000 m, 2000 m, 3000 m, 4000 m, and 5000 m, while the unseen set comprises intermediate distances (1500 m, 2500 m, 3500 m, and 4500 m). The resulting train-validation-test splits for both seen and unseen sets are summarized in Table~\ref{tab:dataset_splits_combined}.

\subsection{Metrics}
We evaluate the fusion models using dataset-specific metrics, as detailed below.

\textbf{Waymo Open Dataset:} We evaluate the performance of the fused networks on the Waymo dataset using 3D mean Average Precision (mAP) and mean Average Precision with Heading (mAPH). Evaluations were conducted for three object classes---Vehicle, Pedestrian, and Cyclist---at IoU thresholds of 0.7, 0.5, and 0.5, respectively. Performance was reported across two difficulty levels (L1 and L2), which are defined in the dataset itself based on the number of LiDAR points associated with each object.

\textbf{ATR Dataset:} We evaluate the fusion network using standard metrics: mean Average Precision across IoU thresholds from 0.50 to 0.95 ($mAP$), alongside mean Average Precision at IoU thresholds of 0.50 and 0.75 ($mAP_{50}$, $mAP_{75}$), and mean Average Recall at 100 proposals ($mAR_{100}$). We report the combined overall performance across both the seen and unseen test splits.

\begin{figure*}[t]
    \centering
    \includegraphics[width=\linewidth]{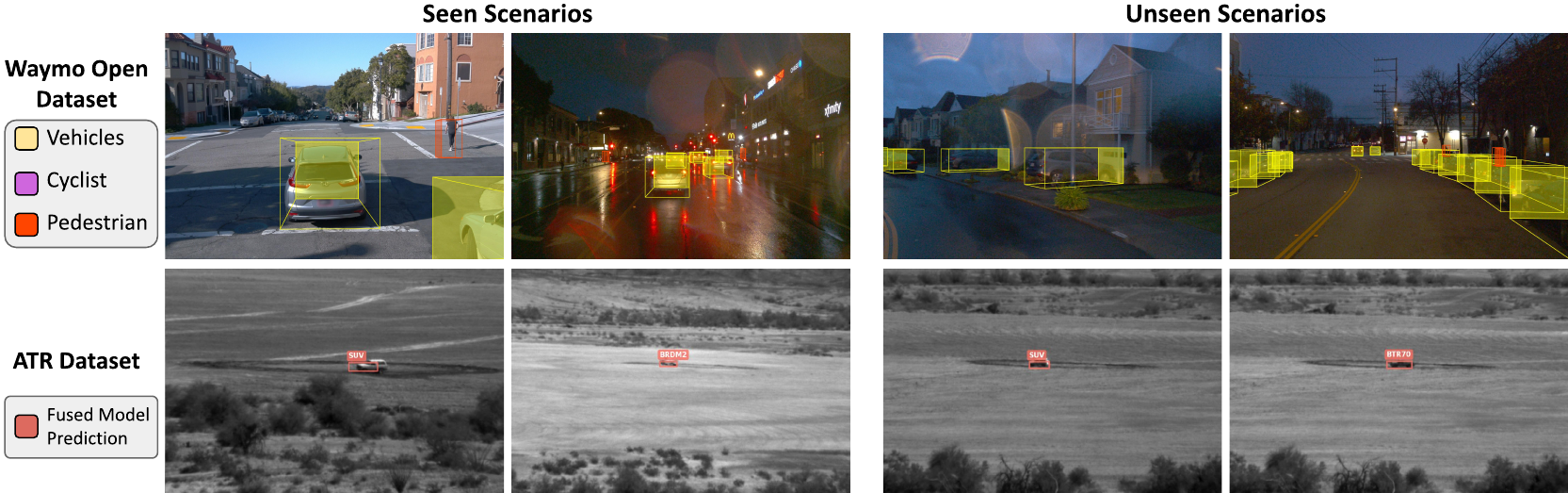}
    \caption{Qualitative examples of VLC Fusion in both seen and unseen environments. \textbf{Top row:} 3D detections on the Waymo Open dataset under \emph{seen} (daytime/nighttime) and \emph{unseen} (dawn/dusk time) conditions, with vehicles (yellow), cyclists (purple), and pedestrians (red) accurately localized. \textbf{Bottom row:} 2D detection on the ATR dataset for \emph{seen} (1000 m and 2000 m distances respectively) and \emph{unseen} (1500 m distance) scenarios. More qualitative examples are provided in Supplementary.}
    \label{fig:sample_predictions}
\end{figure*}

\subsection{Implementation Details}
Below, we describe training setups for individual object detection models per dataset and modality, followed by the generation of environmental conditions using a VLM.

\subsubsection{Object Detectors.}

\textbf{Waymo Open Dataset}: For the RGB modality, we train a DETR-based 2D object detector with a ResNet‑50 backbone using the Hugging Face Trainer. The model was trained for 150 epochs with a batch size of 16 and an initial learning rate of $5 \times 10^{-5}$. Best checkpoints were saved every 250 steps using validation mAP.

For the LiDAR modality, we used the SECOND 3D object detection model, trained on Waymo point-cloud data. Training followed the standard MMDetection3D pipeline, with the point-cloud range set from $[-76.8, -51.2, -2]$ m to $[76.8, 51.2, 4]$ m, targeting the Car, Pedestrian, and Cyclist classes. The model was trained for 100 epochs using the AdamW optimizer with a learning rate of $1 \times 10^{-3}$ and a batch size of 2. Evaluation was conducted after each epoch using the \texttt{WaymoMetric} evaluator, and the best-performing checkpoint was selected for downstream use. Additionally, each fusion model was fine-tuned for 40 epochs using the same optimization settings as the LiDAR detector.

\textbf{ATR Dataset}: We train two separate DETR-based 2D object detectors with a ResNet‑50 backbone, one for visible images and the other for MWIR images, using the Hugging Face Trainer. Each model was trained for 140 epochs using the AdamW optimizer with an initial learning rate of $5 \times 10^{-5}$ and a weight decay of $1 \times 10^{-4}$. Training used a batch size of 32 with gradient accumulation over 8 steps. Model checkpoints were evaluated based on validation mean Average Precision (mAP), and the checkpoint achieving the highest mAP was retained for final evaluation. Additionally, for the ATR dataset, each fusion model was fine-tuned for 100 epochs using the same training configuration.

\subsubsection{VLM-queried Environmental Conditions.}
We first generated environmental conditions using the methods described in Section~\ref{sec:condition_extraction}. Two sets of conditions were defined: human-defined and automatically extracted. After obtaining these conditions, we queried GPT-4o to generate responses for each data point.

\begin{table}[t]
    \caption{Overall Waymo performance on seen and unseen scenarios. We report 3D mAP/mAPH for both L1 and L2 difficulty levels. Best and second-best results are highlighted in bold and underline, respectively.}
    \centering
    \setlength{\tabcolsep}{4pt}
    \scriptsize
    \resizebox{\linewidth}{!}{%
    \begin{tabular}{l|cc|cc}
        \toprule
        \multirow{2}{*}{\textbf{Fusion Technique}} &
        \multicolumn{2}{c|}{\textbf{Seen Scenario (mAP/mAPH)}} &
        \multicolumn{2}{c}{\textbf{Unseen Scenario (mAP/mAPH)}} \\
        \cmidrule(lr){2-3} \cmidrule(lr){4-5}
        & \textbf{L1} & \textbf{L2}
        & \textbf{L1} & \textbf{L2} \\
        \midrule
        Fusion SSD
        & 26.1/23.6 & 23.06/20.8
        & 31.03/28.08 & 28.6/25.8 \\

        \makecell[l]{Fusion SSD with\\Self-Attention}
        & 21.8/19.6 & 19.07/17.1
        & 24.2/21.8 & 22.2/19.9 \\

        Learnable Align
        & 18.6/16.3 & 16.2/14.2
        & 23.08/20.1 & 21.2/18.5 \\

        RGB-X
        & {27.1/24.5} & {23.9/21.6}
        & {31.03/28.1} & {28.5/25.9} \\

        \midrule
        \makecell[l]{VLC Fusion with Human\\Defined Conditions (n=3)}
        & \underline{28.5/26.2} & \underline{25.2/23.08}
        & \underline{32.3/29.7} & \underline{29.8/27.3} \\

        \makecell[l]{VLC Fusion with Extracted\\Conditions (n=10)}
        & \textbf{37.0/34.4} & \textbf{33.0/30.5}
        & \textbf{41.5/38.3} & \textbf{38.4/35.4} \\
        \bottomrule
    \end{tabular}}
    \label{tab:waymo_overall_seen_unseen}
\end{table}

\begin{figure}[t]
    \centering
    \begin{subfigure}[t]{0.48\linewidth}
        \centering
        \includegraphics[width=0.85\linewidth]{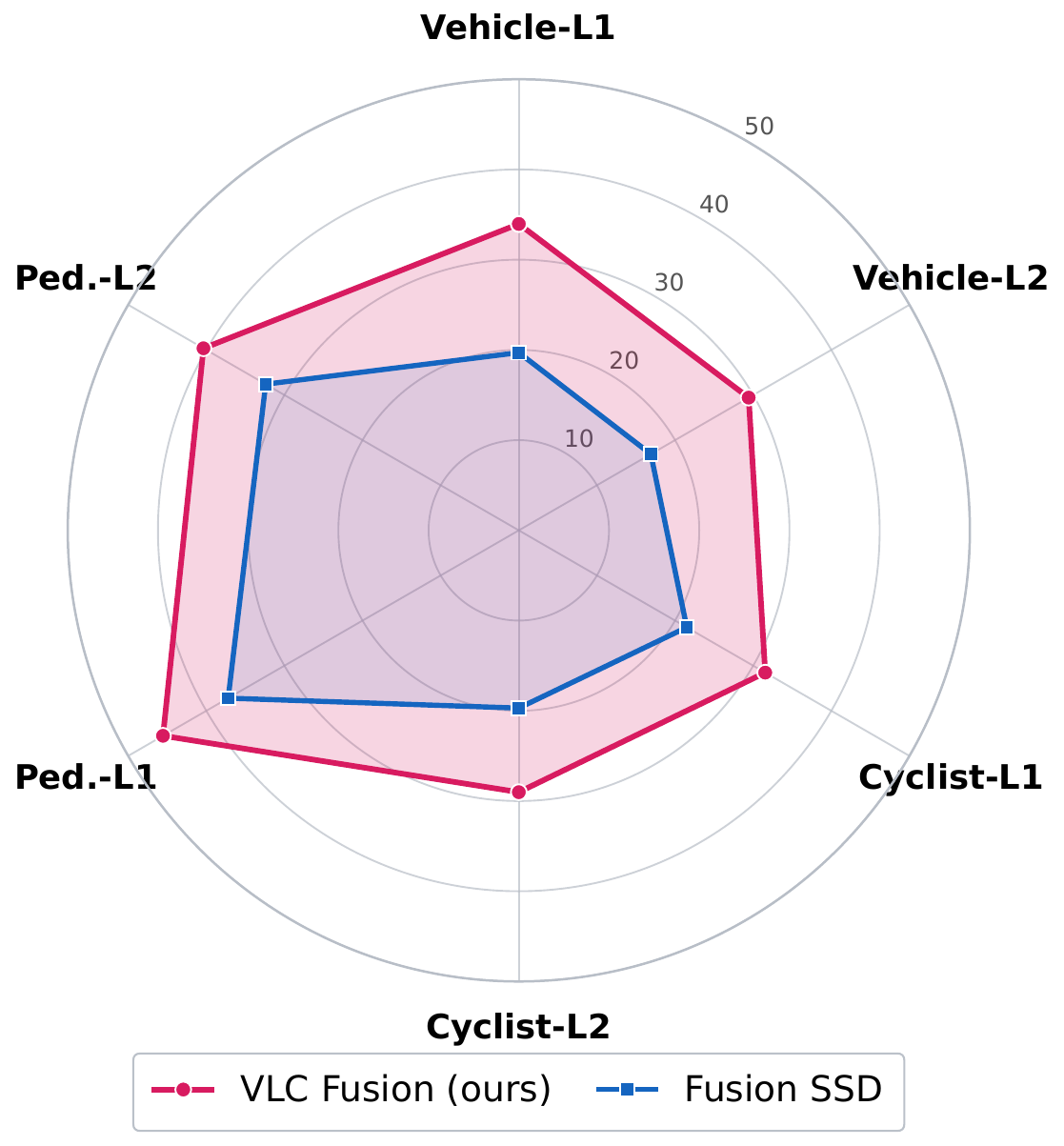}
        \caption{Seen scenario}
        \label{fig:waymo_seen_radar}
    \end{subfigure}
    \hfill
    \begin{subfigure}[t]{0.48\linewidth}
        \centering
        \includegraphics[width=0.85\linewidth]{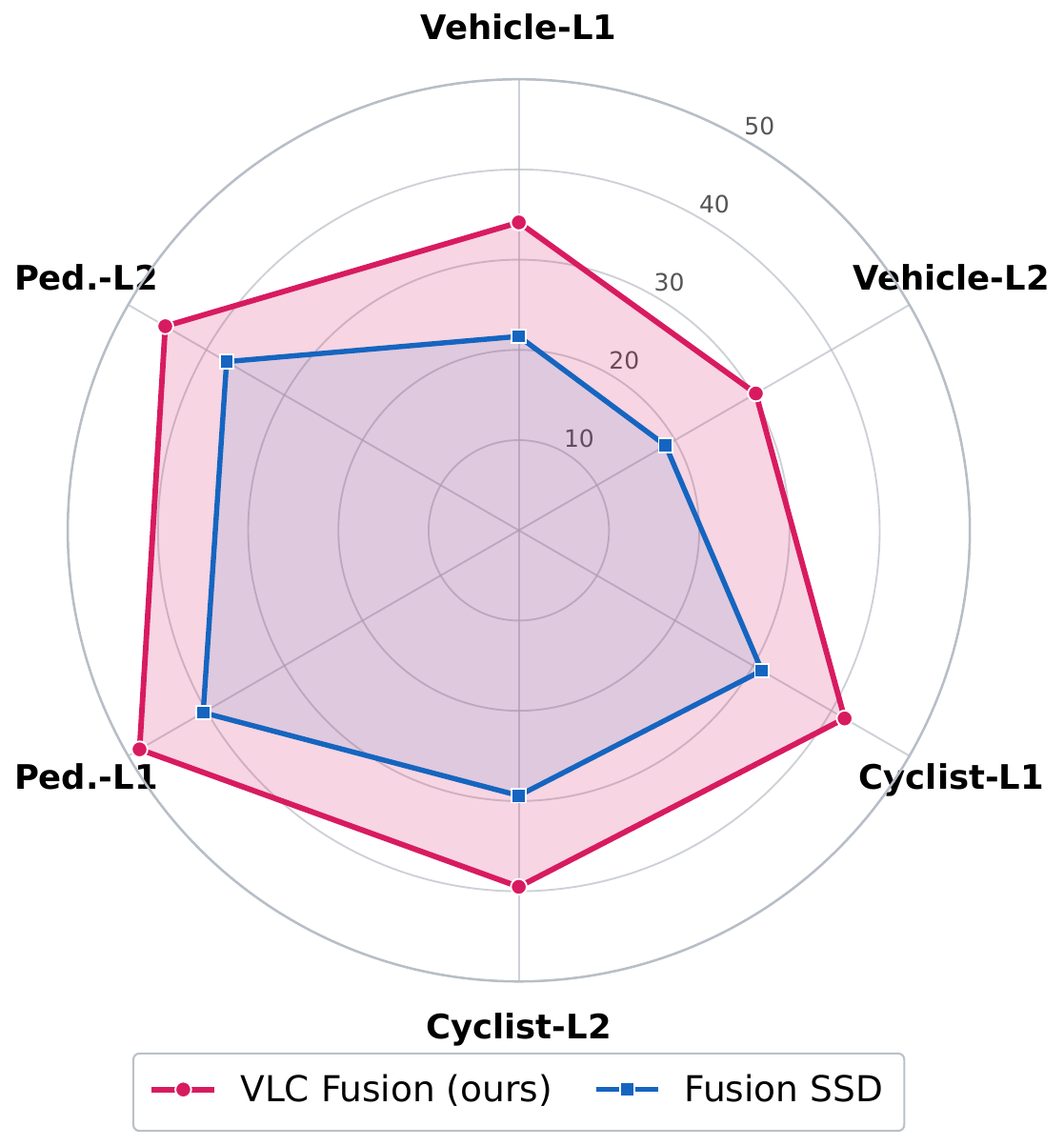}
        \caption{Unseen scenario}
        \label{fig:waymo_unseen_radar}
    \end{subfigure}
    \caption{Per-class 3D AP (\%) comparison between VLC Fusion and Fusion SSD on the Waymo dataset under (a) seen and (b) unseen scenarios.}
    \label{fig:waymo_seen_unseen_vlc_vs_fusionssd}
\end{figure}

\subsection{Baselines}
We explored various fusion strategies, including Fusion SSD~\cite{bijelic2020seeing}, Fusion SSD with self-attention, RGB-X~\cite{deevi2024rgb}, and Learnable Align~\cite{li2022deepfusion}:

\textbf{Fusion SSD and Variations:} The base Fusion SSD architecture concatenates feature maps from both modalities and applies convolution to reduce them to the appropriate dimensions before passing them to the detection head. In the self-attention variant, an additional attention module is applied after the convolution step to re-weight features based on their importance. 

\textbf{RGB-X:} 
In this approach, feature maps are concatenated and processed using CBAM, which applies channel attention followed by spatial attention to refine the fused representation. This two-step attention mechanism adaptively emphasizes both the most informative channels and spatial regions. The refined feature map is then passed through convolutional layers to align its dimensionality with the detection head.

\textbf{Learnable Align:} We also evaluated Learnable Align, where a lightweight cross-attention block is used to fuse features from the two modalities. In this method, each spatial cell in one modality’s feature map is treated as a query, while the corresponding features from the other modality serve as keys and values. This end-to-end attention mechanism enables the model to align and highlight the most relevant information across modalities.

Each baseline fusion method was trained using the standard detection head for its dataset: a DETR‐based 2D head for ATR and a SECOND‐based 3D head for Waymo. All baseline methods were trained/evaluated without environmental conditions.

\subsection{Results}

\begin{table}[t]
    \caption{Overall ATR performance on seen and unseen test splits. We report $mAP_{50}$, $mAP_{75}$, overall $mAP$ (computed over IoU thresholds from 0.50 to 0.95 with a step size of 0.05), and $mAR_{100}$. Best and second-best results are highlighted in bold and underline, respectively.}
    \centering
    \setlength{\tabcolsep}{3.5pt}
    \scriptsize
    \resizebox{\linewidth}{!}{%
    \begin{tabular}{l|cccc|cccc}
        \toprule
        \multirow{2}{*}{\textbf{Fusion Technique}} &
        \multicolumn{4}{c|}{\textbf{Seen Scenario}} &
        \multicolumn{4}{c}{\textbf{Unseen Scenario}} \\
        \cmidrule(lr){2-5} \cmidrule(lr){6-9}
        & \textbf{$mAP_{50}$} & \textbf{$mAP_{75}$} & \textbf{$mAP$} & \textbf{$mAR_{100}$}
        & \textbf{$mAP_{50}$} & \textbf{$mAP_{75}$} & \textbf{$mAP$} & \textbf{$mAR_{100}$} \\
        \midrule
        Fusion SSD
        & \underline{87.19} & \underline{67.79} & \underline{57.82} & 64.69
        & 23.59 & \underline{5.56} & \underline{9.42} & \underline{14.94} \\

        \makecell[l]{Fusion SSD w/\\Self-Attention}
        & 82.82 & 57.79 & 51.12 & 57.40
        & 22.01 & 4.07 & 8.29 & 13.30 \\

        RGB-X
        & 82.04 & 58.48 & 51.60 & 57.45
        & 20.67 & 5.28 & 8.43 & 12.88 \\

        Learnable Align
        & 86.97 & 67.14 & 57.62 & \underline{64.84}
        & 15.57 & 3.80 & 6.21 & 10.96 \\

        \midrule
        \makecell[l]{VLC Fusion with Human\\ Conditions (n=3)}
        & 84.07 & 56.35 & 51.39 & 57.53
        & \underline{23.65} & 1.68 & 7.46 & 11.93 \\

        \makecell[l]{VLC Fusion with Extracted\\ Conditions (n=7)}
        & \textbf{96.41} & \textbf{69.37} & \textbf{61.67} & \textbf{81.95}
        & \textbf{32.22} & \textbf{9.48} & \textbf{13.38} & \textbf{35.51} \\
        \bottomrule
    \end{tabular}}
    \label{tab:atr_overall_seen_unseen}
\end{table}

\begin{figure}[t]
    \centering
    \begin{subfigure}[t]{0.48\linewidth}
        \centering
        \includegraphics[width=0.85\linewidth]{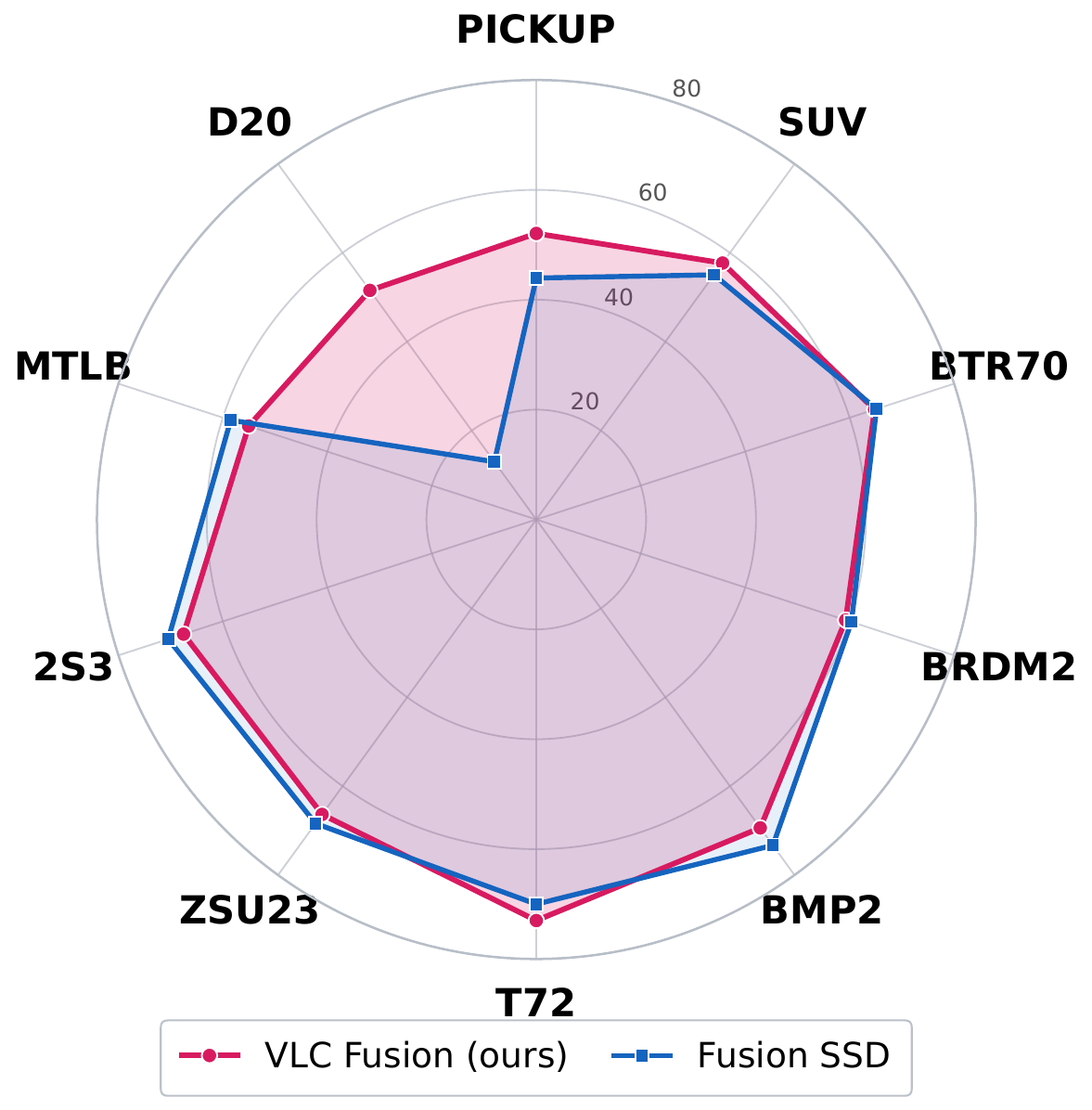}
        \caption{Seen scenario}
        \label{fig:atr_seen_radar}
    \end{subfigure}
    \hfill
    \begin{subfigure}[t]{0.48\linewidth}
        \centering
        \includegraphics[width=0.85\linewidth]{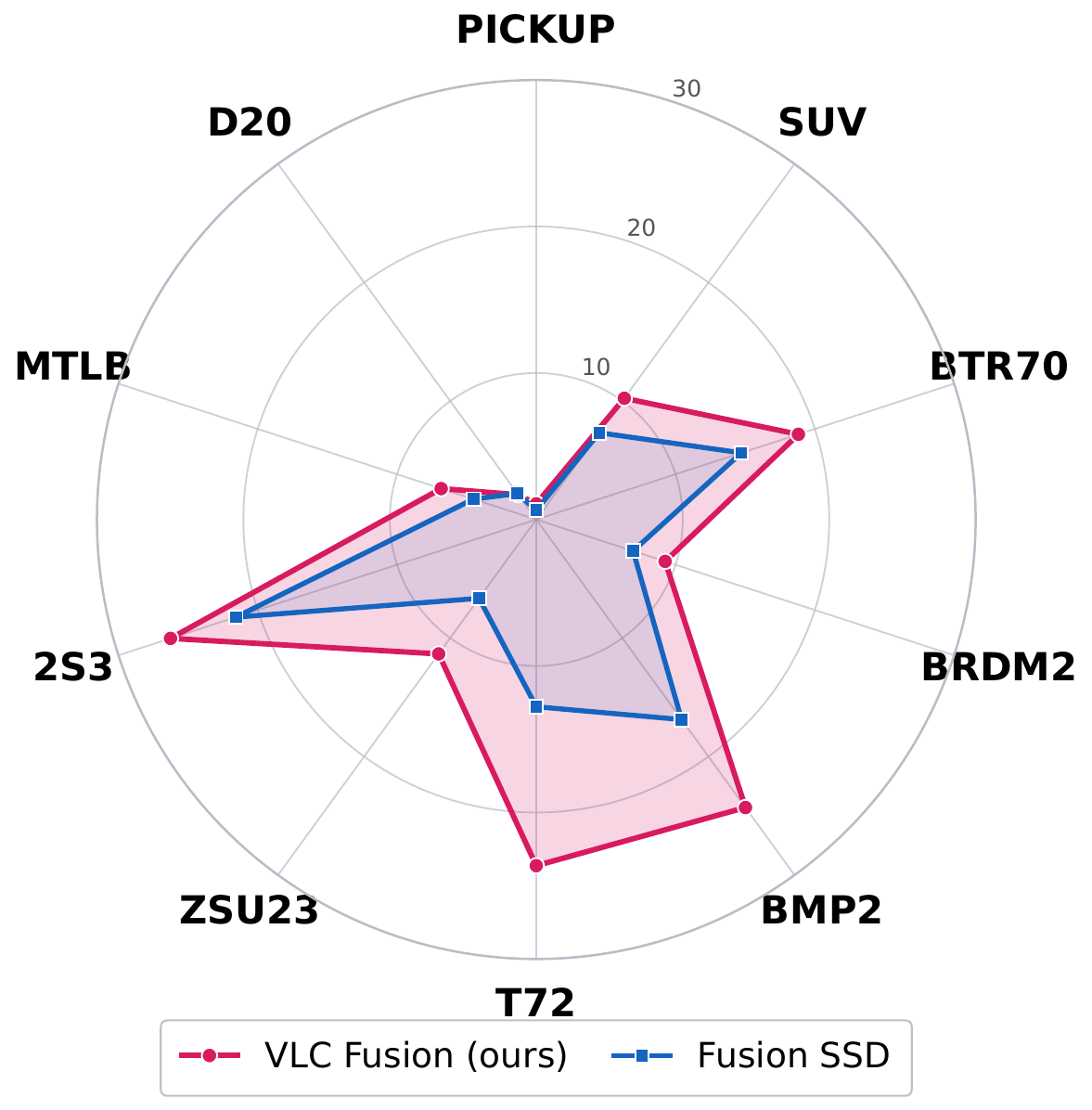}
        \caption{Unseen scenario}
        \label{fig:atr_unseen_radar}
    \end{subfigure}
    \caption{Per-class AP (\%) comparison between VLC Fusion and Fusion SSD on the ATR dataset under (a) seen and (b) unseen scenarios.}
    \label{fig:atr_seen_unseen_vlc_vs_fusionssd}
\end{figure}

We evaluate whether environmental conditions improve multimodal sensor fusion performance, using test sets from Table~\ref{tab:dataset_splits_combined}.

\textbf{Waymo Open Dataset:} Tables~\ref{tab:waymo_overall_seen_unseen} and Fig.~\ref{fig:waymo_seen_unseen_vlc_vs_fusionssd} show that VLC Fusion with 10 conditions consistently outperforms other baseline algorithms. Specifically, VLC Fusion achieved a 3D mAP of 37.0\% in the Day and Night (seen) scenario and 41.5\% in the Dawn/Dusk (unseen) scenario.
Interestingly, increasing the number of conditions from 3 to 10 notably improved accuracy for underrepresented classes such as cyclists by 10.8\% and 14.3\% in the seen and unseen scenarios, respectively. This suggests that incorporating additional environmental context helps the fusion model generalize to challenging, underrepresented scenarios. Qualitative examples from Fig.~\ref{fig:sample_predictions} further support these findings, illustrating VLC Fusion’s enhanced capability to detect vehicles, cyclists, and pedestrians under varied environmental conditions.

\textbf{ATR Dataset:} 
Tables~\ref{tab:atr_overall_seen_unseen} and Fig.~\ref{fig:atr_seen_unseen_vlc_vs_fusionssd} reinforce the effectiveness of VLC Fusion. Despite the ATR dataset presenting comparatively simpler environmental variations, VLC Fusion consistently improved performance across both seen and unseen scenarios. VLC Fusion with 7 VLM-extracted environmental conditions performs best on the seen and unseen test splits, achieving a mAP of 61.67\% and 13.38\%, surpassing Fusion SSD’s best baseline result of 57.82\% and 9.42\%, respectively. These results emphasize that even datasets with less pronounced environmental variation benefit from incorporating context-specific environmental conditions into the fusion model. Additional results with per-class $mAR_{100}$ are provided in Supplementary.

\begin{figure}[t]
    \begin{minipage}{0.55\textwidth}
        \centering
          \label{tab:inference_time}
          \resizebox{\linewidth}{!}{\begin{tabular}{@{}lclc@{}}
            \toprule
            \textbf{Model} & \textbf{Param} & \textbf{Time} & \textbf{Condition} \\
            & \textbf{in Billions} & \textbf{(sec/image)} & \textbf{Accuracy} \\
            \midrule
            GPT-4o (baseline)$^\dagger$ & $>$ 100 & 2 sec & 100\% \\
            Moondream2 & 1.9 & 0.7 sec & 79.8\% \\
            SmolVLM-Instruct & 2.2 & 1 sec & 56.3\% \\
            \bottomrule
          \end{tabular}}
      {\scriptsize $^\dagger$Parameter count is based on public estimates.}
        \label{table:}
    \end{minipage} \hfill
    \begin{minipage}{0.4\textwidth}
        \centering
        \includegraphics[width=\textwidth]{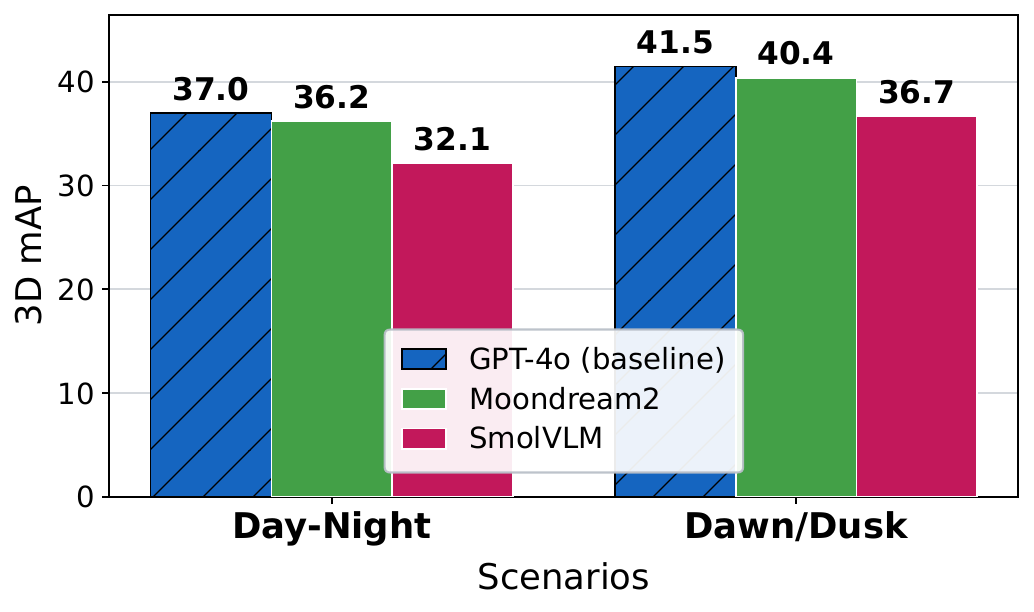}
        \label{fig:transference_abalation}
    \end{minipage}
    \caption{\textbf{(a)} As shown in the table, smaller VLMs, such as Moondream2 and SmolVLM-Instruct, deliver faster inference, but yield lower condition-generation accuracy compared to GPT-4o. (Here the accuracy is calculated by considering the response from GPT-4o as ground truth.) \textbf{(b)} In zero-shot testing, the corresponding drop in 3D mAP (L1 difficulty) on the Waymo Day-Night and Dawn/Dusk scenarios highlights the importance of condition quality.}
    \label{fig:small_vlm_performance}
\end{figure}

\subsection{Ablation Study}
To better understand the role of VLM-based environmental conditions in VLC Fusion, we conduct ablations on four factors: (i) the scale of the VLM used for querying conditions, (ii) the number and consistency of queried conditions, (iii) the semantic alignment of generated captions using CLIP similarity, and (iv) the depth of the VLC Fusion module.

\begin{figure}[t]
    \centering
    \begin{subfigure}[t]{0.48\linewidth}
        \centering
        \includegraphics[width=\linewidth]{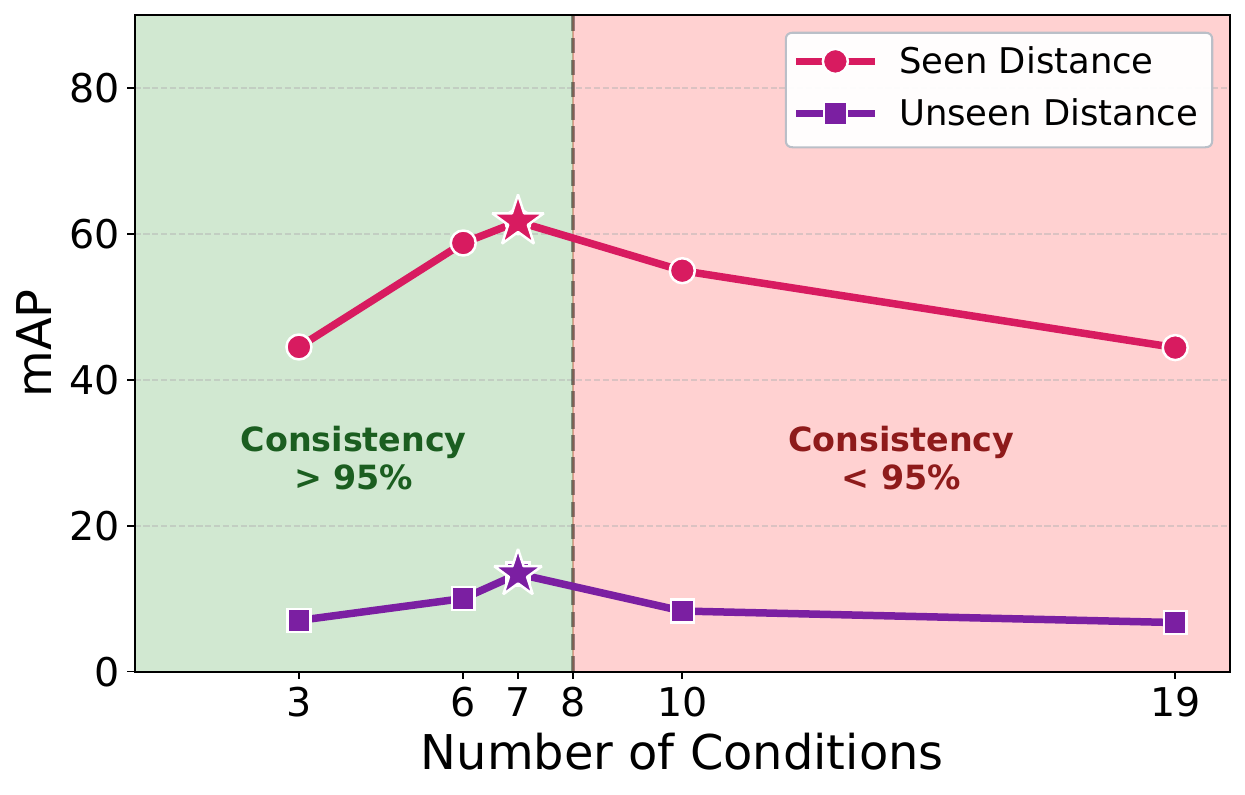}
        \caption{ATR dataset. Performance peaks at 7 conditions, then declines as less consistent conditions are included.
        }
        \label{fig:atr_condition_vs_map}
    \end{subfigure}
    \hfill
    \begin{subfigure}[t]{0.48\linewidth}
        \centering
        \includegraphics[width=\linewidth]{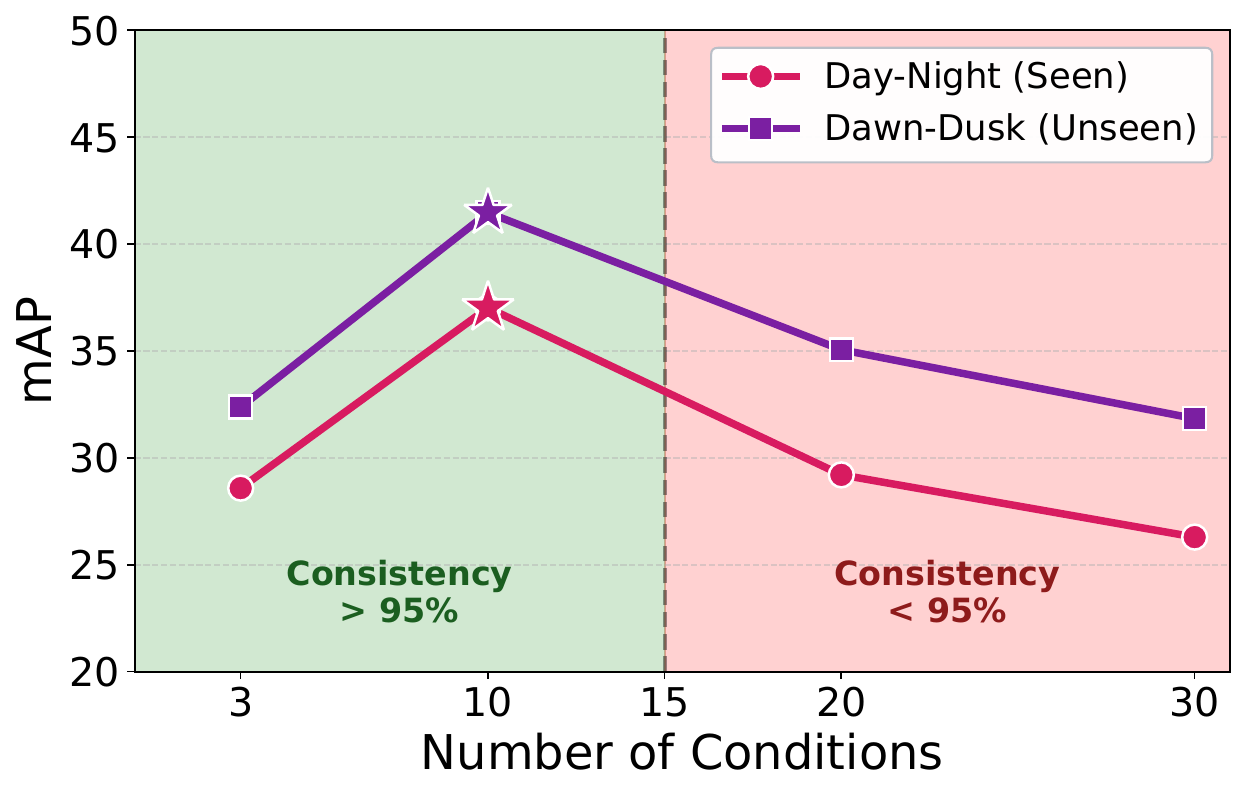}
        \caption{Waymo dataset. Performance peaks at 10 conditions, then declines as less consistent conditions are included.
        }
        \label{fig:waymo_condition_vs_map}
    \end{subfigure}
    \vspace{1em} 
    
    \begin{subfigure}[t]{0.48\linewidth}
        \centering
        \includegraphics[width=\linewidth]{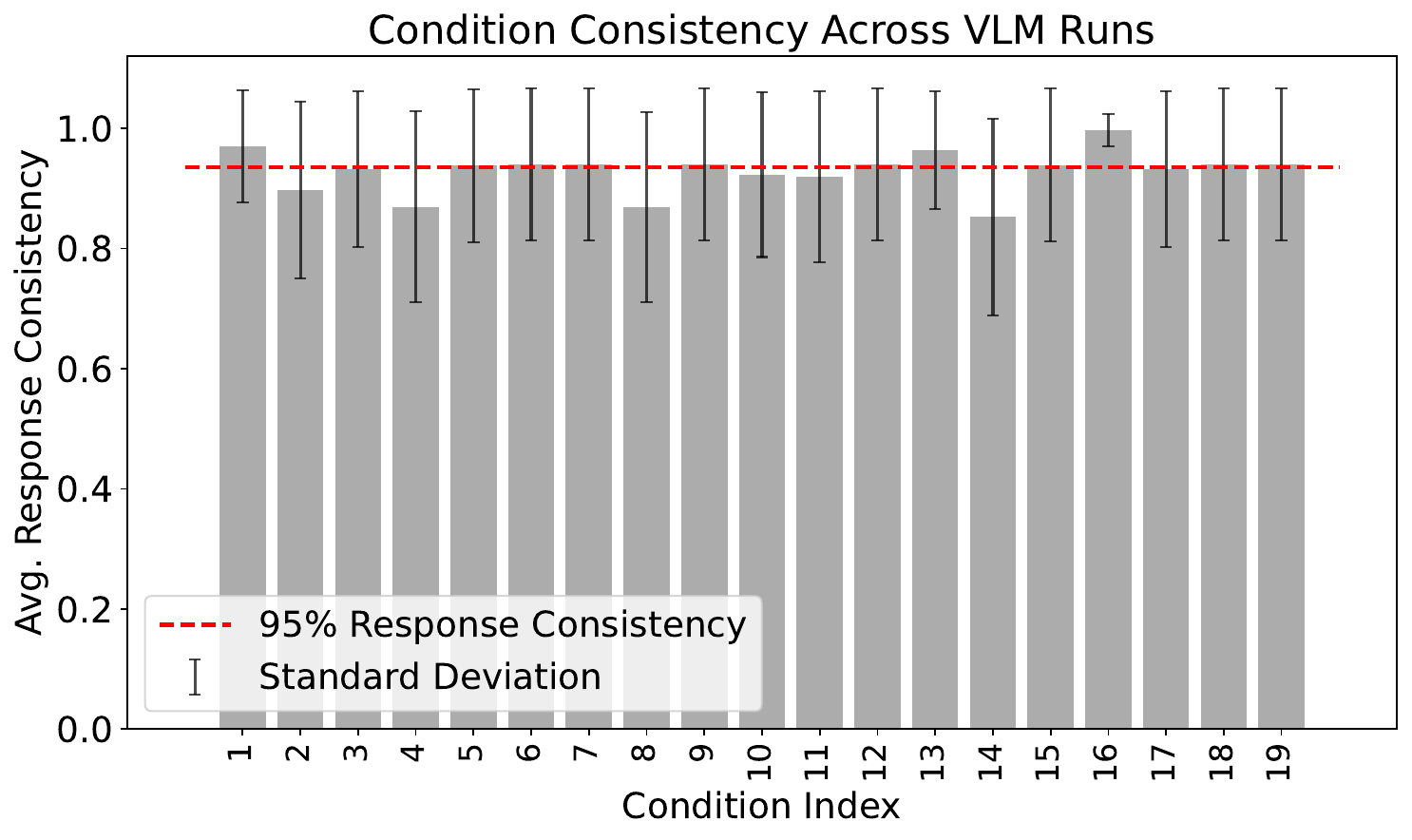}
        \caption{Average condition response consistency over 5 runs for the ATR dataset.}
        \label{fig:atr_condition_consistency}
    \end{subfigure}
    \hfill
    \begin{subfigure}[t]{0.48\linewidth}
        \centering
        \includegraphics[width=\linewidth]{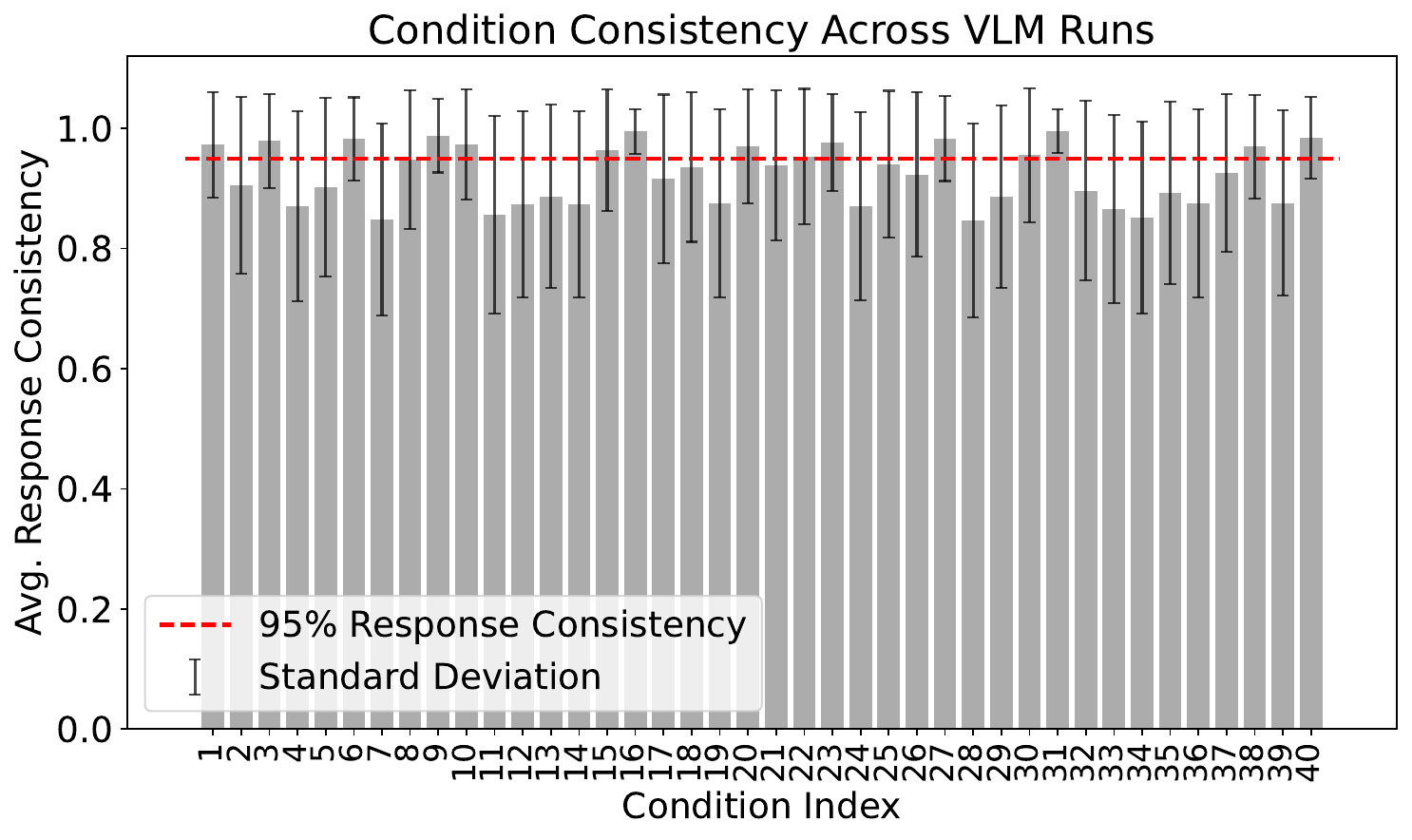}
        \caption{Average condition response consistency over 5 runs for the Waymo dataset.}
        \label{fig:waymo_condition_consistency}
    \end{subfigure}
    
    \caption{Performance of VLC Fusion and corresponding condition consistency. Top row: Performance as a function of the number of ordered environmental conditions. In both datasets, performance initially improves as more conditions are added, but eventually declines when lower-consistency conditions are included. Bottom row: Average condition response consistency across 5 runs.}
    \label{fig:condition_vs_map}
\end{figure}

\subsubsection{Effect of Using Small-scale VLMs for Querying Conditions.}
In this section, we investigate how the scale and capacity of the Vision-Language Model (VLM) used for querying environmental conditions affect detection performance. Intuitively, we expect larger-scale VLMs to produce more accurate and semantically richer environmental condition predictions, thus enhancing the performance of the fused network. Conversely, smaller-scale VLMs are more practical but provide limited semantic reasoning capabilities and less accurate condition predictions, thus potentially reducing fusion performance.

We compared two smaller-scale VLMs (Moondream2~\cite{vik_2024} and SmolVLM~\cite{marafioti2025smolvlm}) against larger-scale VLM (GPT-4o). As shown in Fig.~\ref{fig:small_vlm_performance}, use of small VLMs slightly reduced the performance compared to the GPT-4o. Specifically, the performance for the ``Day-Night (seen)'' scenario drops from 37.0 to 36.2 with Moondream2 and further to 32.1 with SmolVLM. Similarly, for the ``Dawn-Dusk (unseen)'' scenario, performance decreases from 41.5 to 40.4 (Moondream2) and 36.7 (SmolVLM). These results confirm our hypothesis that the scale of VLM influences the accuracy of environmental condition predictions and the overall performance of VLC Fusion. On a positive note, Moondream2 remains competitive with GPT-4o, suggesting that smaller VLMs can offer a practical trade-off between efficiency and accuracy.

\subsubsection{Effect of Condition Quantity and Consistency.}
We further analyze how varying the number and consistency of queried environmental conditions impacts the fusion model's performance. Fig.~\ref{fig:atr_condition_vs_map} and Fig.~\ref{fig:waymo_condition_vs_map} show the trend observed in our experiments. Initially, increasing the number of conditions leads to improvement in detection performance. However, beyond a certain point, adding more conditions leads to a performance decline. This trend is attributed to incorporating less consistent and potentially noisy conditions. As shown in Fig.~\ref{fig:atr_condition_consistency} and~\ref{fig:waymo_condition_consistency}, we observed that conditions ranked higher in consistency contributed positively to performance, whereas adding less consistent conditions diminished model accuracy.
Thus, our results underscore the critical importance of selecting a carefully curated set of highly consistent conditions, balancing richness of contextual information with the risk of introducing noise or irrelevant context.

\begin{figure}[t]
    \centering

    \begin{minipage}[t]{0.48\linewidth}
        \vspace{0pt}
        \centering
        \captionof{table}{Average CLIP-based cosine similarity between images and generated captions. Both the Waymo and ATR datasets achieve scores within the optimal 0.3 to 0.4 threshold for open-world tasks, demonstrating strong semantic alignment of the extracted conditions.}
        \label{tab:clip_similarity}
        \scriptsize
        \setlength{\tabcolsep}{12pt}
        \begin{tabular}{lc}
            \toprule
            \textbf{Dataset} & \textbf{CLIP Score} \\
            \midrule
            Waymo Dataset & 0.314 $\pm$ 0.020 \\
            ATR Dataset   & 0.335 $\pm$ 0.023 \\
            \bottomrule
        \end{tabular}
    \end{minipage}
    \hfill
    \begin{minipage}[t]{0.49\linewidth}
        \vspace{0pt}
        \centering
        \includegraphics[width=\linewidth]{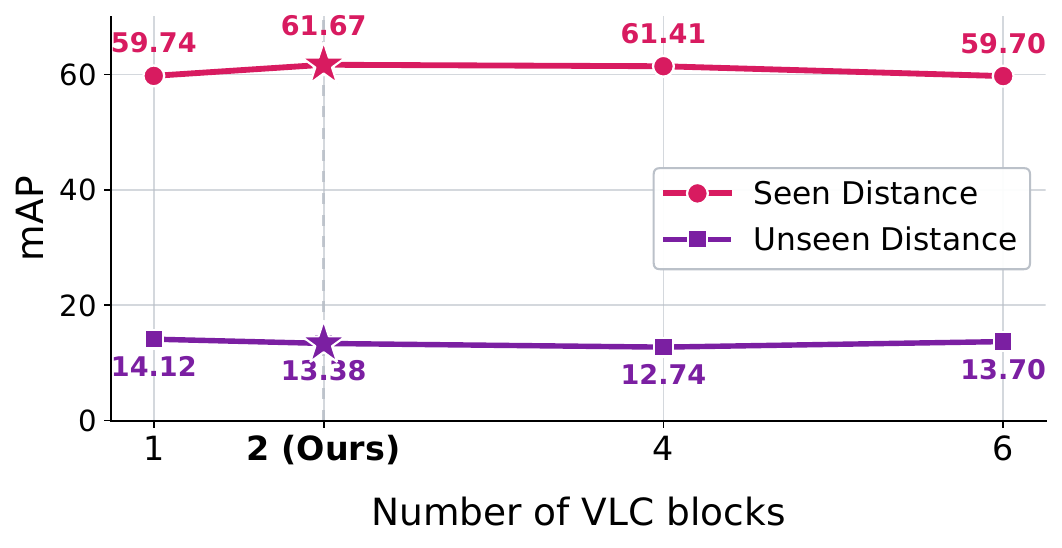}
        \captionof{figure}{Impact of number of VLC blocks on overall performance for the ATR dataset. mAP peaks at 2 stacked blocks before degrading.}
        \label{fig:atr_depth_ablation}
    \end{minipage}
\end{figure}

\subsubsection{CLIP-Based Caption Quality Analysis.}
To quantitatively assess the alignment between generated captions and visual content, we compute the cosine similarity between image and caption embeddings using CLIP (ViT-B/32)~\cite{radford2021learning}. As detailed in~\cite{hessel2021clipscore}, CLIP similarity scores between 0.3 and 0.4 indicate good alignment in open-world captioning tasks. As shown in Table~\ref{tab:clip_similarity}, our results indicate that the generated captions are semantically coherent with the input images, supporting their reliability in condition generation for VLC Fusion.

\subsubsection{Effect of VLC Fusion Block Depth.}
To determine the optimal architecture for our fusion module, we further evaluate how the number of stacked VLC blocks (network depth) influences detection accuracy on the ATR dataset. As shown in Fig.~\ref{fig:atr_depth_ablation}, increasing the network depth initially improves performance by allowing the model to learn more complex multimodal interactions. However, performance peaks at 2 VLC blocks, achieving a mAP of 61.67 for seen distances. Adding further blocks beyond this optimal depth leads to performance degradation, with accuracy dropping to 59.70 at 6 blocks, likely due to overfitting. Therefore, we utilize 2 VLC blocks as the default configuration for our experiments.

\section{Conclusion}

In this paper, we introduced Vision-Language Conditioned Fusion (VLC Fusion), a novel sensor fusion framework that improves object detection robustness by dynamically conditioning multimodal fusion on environmental context queried from VLMs. Unlike conventional fusion methods that use fixed fusion strategies, VLC Fusion explicitly incorporates high-level semantic information about the environment, enabling the model to better adapt sensor weighting under diverse and previously unseen conditions.

We evaluated VLC Fusion on two real-world multimodal datasets: the Waymo Open dataset for autonomous driving and the ATR dataset for military target recognition. Across both datasets, VLC Fusion consistently outperformed existing fusion baselines in seen and unseen scenarios, demonstrating its effectiveness for improving both detection accuracy and robustness. Our ablation studies further showed that the quality and consistency of environmental conditions are critical: adding reliable contextual information improves performance, while incorporating excessive or noisy conditions can reduce the benefit.

Overall, our results highlight the potential of integrating VLM-based semantic reasoning into sensor fusion architectures for more reliable perception in complex environments. Future work includes improving automatic condition extraction, extending VLC Fusion to additional sensor modalities and deployment settings, and investigating efficient real-time deployment scenarios.

\begin{credits}
\subsubsection{\ackname}
This research was supported by the Defense Advanced Research Projects Agency (DARPA) under Cooperative Agreement No. HR00112420370 (MCAI). The views expressed in this paper are those of the authors and do not necessarily reflect the official policy or position of the U.S. Military Academy, the U.S. Army, the U.S. Department of Defense, or the U.S. Government. We would also like to thank Caleb Liu for early discussions on fine-tuning object detectors, and Som Sagar for insights into the applications of FiLM.

\subsubsection{\discintname}
The authors have no competing interests to declare that are relevant to the
content of this article.
\end{credits}


\bibliographystyle{splncs04}
\bibliography{mybibliography}

\clearpage
\section*{Appendix}

\setcounter{section}{0}
\renewcommand{\thesection}{\Alph{section}}

\section{Computational Resources}
\label{app:compute}

All experiments were conducted on NVIDIA H100 GPU (96 GB HBM3e) running Ubuntu 22.04, with CUDA 11.8. For training the detection models, we utilized mixed-precision (FP16) via PyTorch’s AMP module to reduce GPU memory usage and accelerate kernel execution. Memory consumption varied depending on the fusion technique and task, reaching a maximum of approximately 40 GB. Training each model took roughly 4 days. Inference was also performed on the same NVIDIA H100 GPU, with a maximum memory usage of around 10 GB.


\section{Automatic Condition Extraction}
\label{app:conditions}
In this section, we provide the details on conditions extracted from both the dataset. We also evaluate the correctness of captions being generated for both dataset in step 1.

\subsection{Waymo Open Dataset}
\subsubsection{Sample Image-caption Pairs.}
\begin{figure}[t]
    \centering
    \includegraphics[width=1\linewidth]{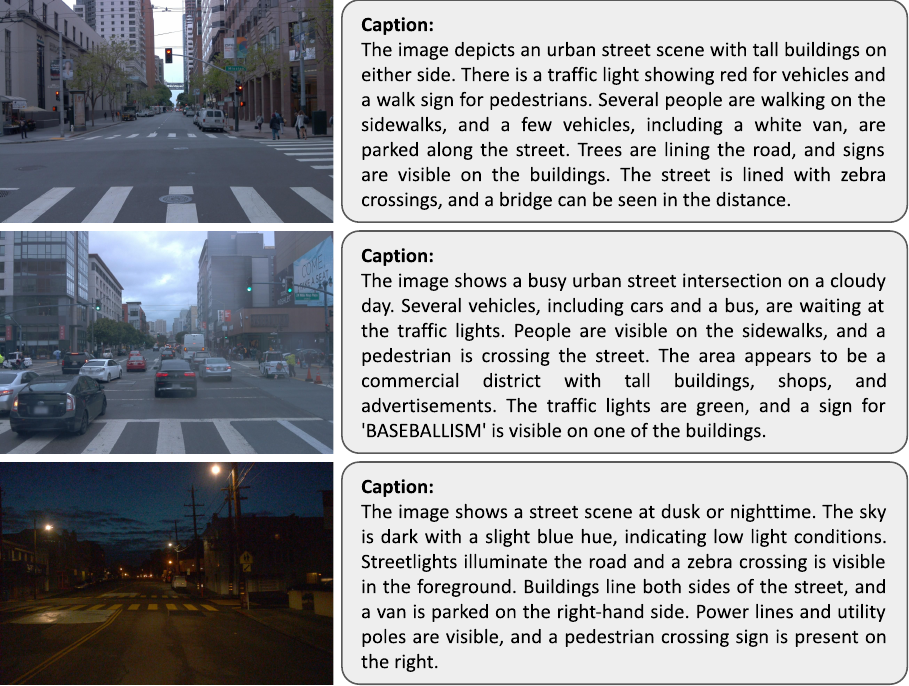}
    \caption{Samples of image-caption pairs generated during automatic condition extraction for Waymo dataset.}
    \label{fig:waymo_image_caption}
\end{figure}

The Fig.~\ref{fig:waymo_image_caption} shown below highlights the sample image-caption pairs created during the automated conditional extraction of Waymo dataset.

\subsubsection{List of Extracted Environmental Conditions.}
Below, we provide the complete list of extracted environmental conditions extracted from Waymo dataset.

\begin{enumerate}
  \item Is the road wet or reflective, possibly due to rain?
  \item Are there any visible pedestrians in the image?
  \item Is there a visible stop sign in the image?
  \item Are there any vehicles parked on the side of the road?
  \item Is a traffic light visible in the image?
  \item Is the image depicting a rainy day?
  \item Are there any tall buildings visible?
  \item Is there a dedicated lane for buses or taxis?
  \item Is the scene set during nighttime?
  \item Is there construction work visible?
  \item Is there a vehicle in motion in the image?
  \item Are street signs or traffic signs visible?
  \item Is there greenery or trees lining the street?
  \item Is there any advertisement or commercial sign visible?
  \item Are there any bicycles or bicycle lanes visible?
  \item Is there a body of water visible?
  \item Are overhead power lines visible?
  \item Is public transportation, like a bus, visible?
  \item Is a visible crosswalk present?
  \item Are there any orange traffic cones visible?
  \item Is the sky clear and blue?
  \item Are the roads cracked or uneven?
  \item Is there a sense of fog or mist in the image?
  \item Is there a notable commercial establishment visible?
  \item Is a noticeable hill or incline visible?
  \item Is the scene from a residential neighborhood?
  \item Is there an indication of a scenic viewpoint?
  \item Is the scene taking place at an intersection?
  \item Are buildings visible in the scene?
  \item Is traffic congestion visible?
  \item Is a pedestrian bridge or crossing present?
  \item Is there traffic light congestion or light signals visible?
  \item Is the street scene located in an urban environment?
  \item Are there multiple lanes on the road?
  \item Is the weather overcast or cloudy?
  \item Are there parked cars visible on the street?
  \item Is there a visible neon or illuminated sign?
  \item Is the image captured from an elevated perspective?
  \item Is the overall atmosphere calm and quiet?
  \item Is there noticeable lens flare or light artifacts in the image?
\end{enumerate}

\subsubsection{Additional Quantitative Analysis.}
\label{app:B_1_3}
\begin{figure}[t]
    \centering
    \includegraphics[width=1\linewidth]{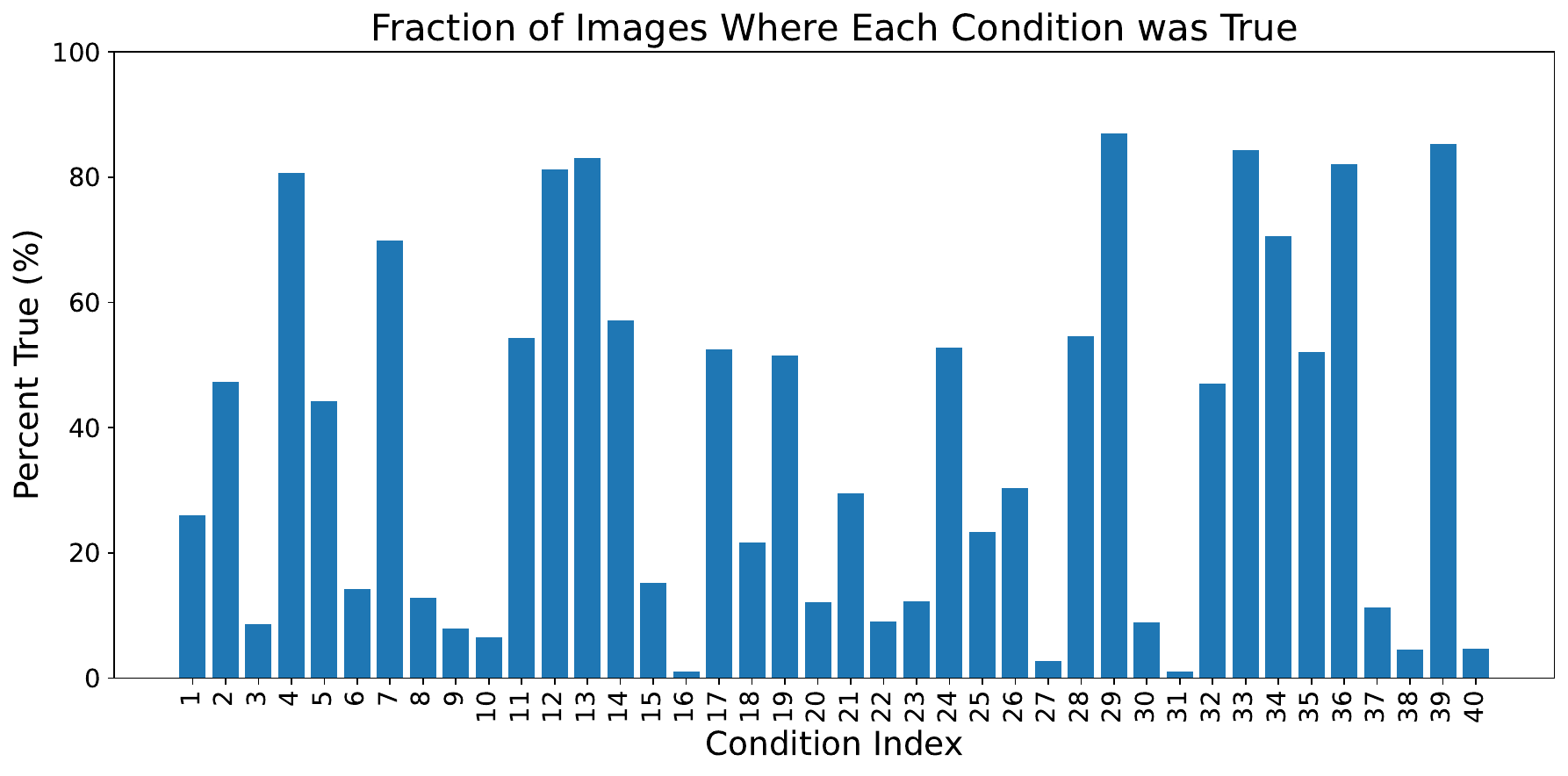}
    \caption{Fraction of images in the Day–Night (seen) test set for which each condition is true in Waymo dataset.}
    \label{fig:waymo_day_night_fraction_true}
\end{figure}

\begin{figure}[t]
    \centering
    \includegraphics[width=1\linewidth]{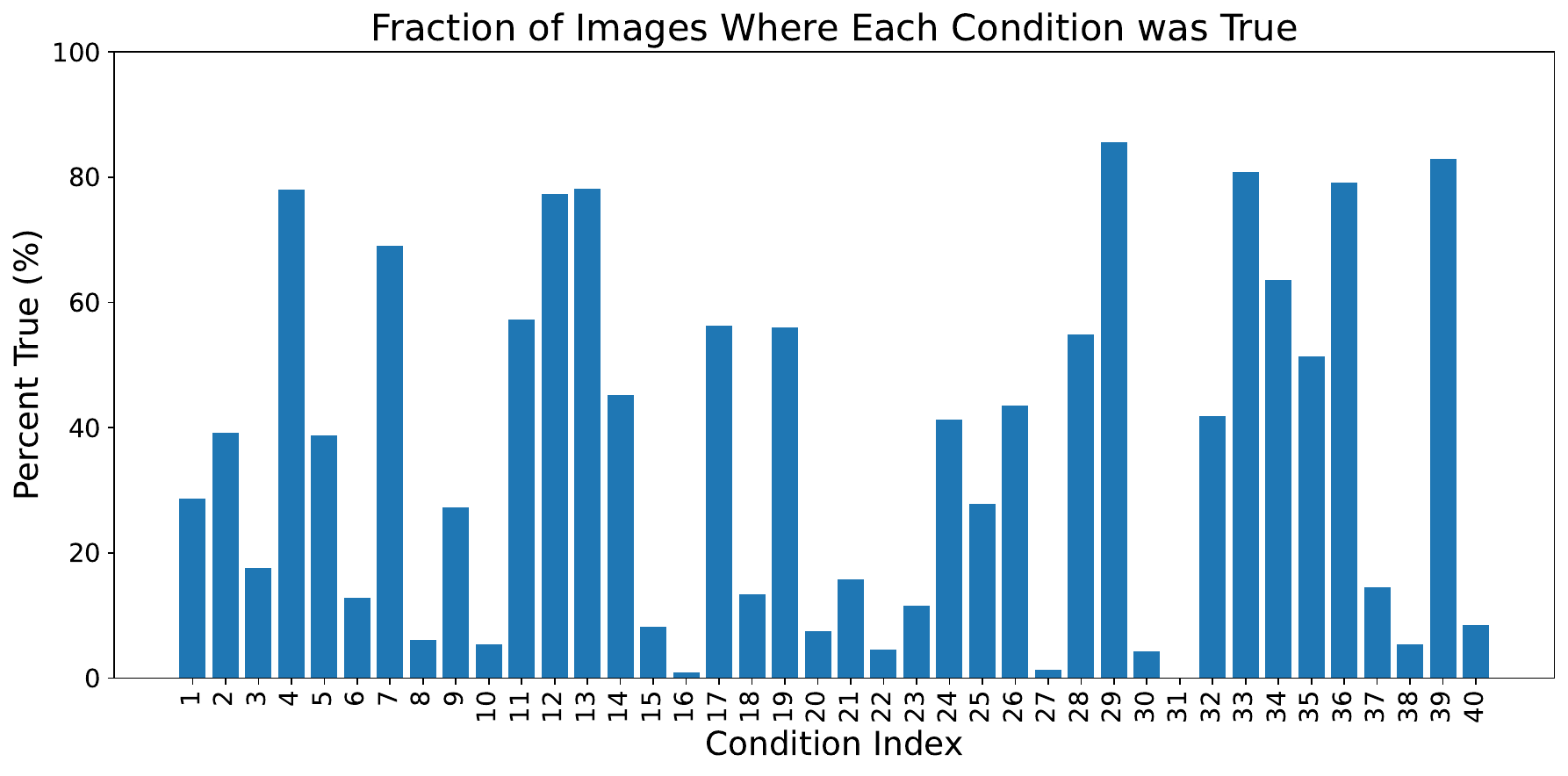}
    \caption{Fraction of images in the Dawn–Dusk (unseen) test set for which each condition is true in Waymo dataset.}
    \label{fig:waymo_dawn_dusk_fraction_true}
\end{figure}

Fig.~\ref{fig:waymo_day_night_fraction_true} and \ref{fig:waymo_dawn_dusk_fraction_true} shows the activation of conditions over Day-night (seen) and Dawn/dusk (unseen) test set. We can see that 85\% of the test dataset have at least one active environmental condition.

\subsection{ATR Dataset}
\subsubsection{Sample Image-caption Pairs.}
\begin{figure}[t]
    \centering
    \includegraphics[width=1\linewidth]{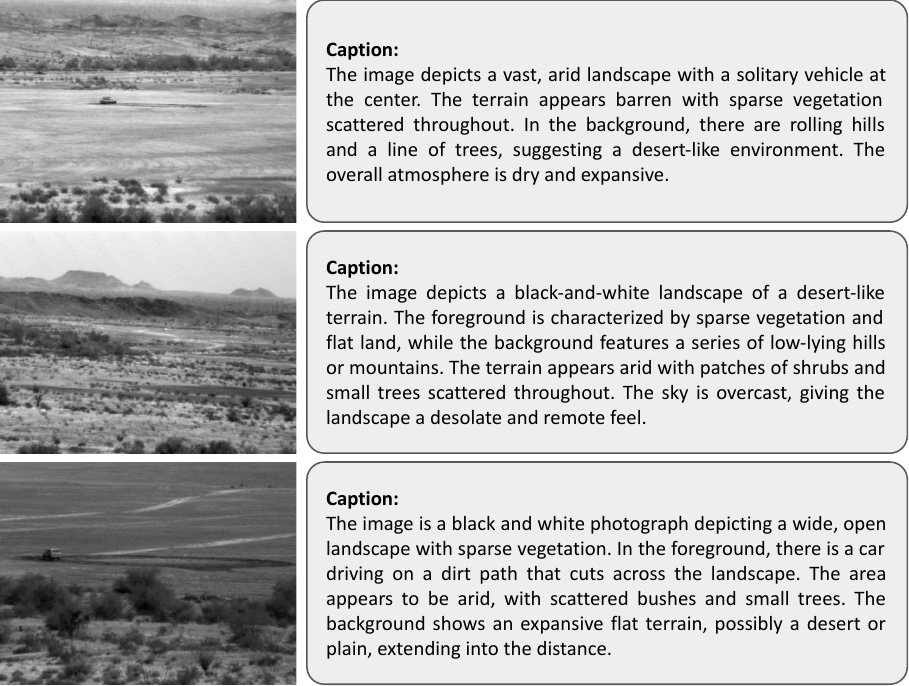}
    \caption{Samples of image-caption pairs generated during automatic condition extraction for ATR dataset.}
    \label{fig:atr_image_caption}
\end{figure}

The Fig.~\ref{fig:atr_image_caption} shown below highlights the sample image-caption pairs created during the automated conditional extraction of ATR dataset.

\subsubsection{List of Extracted Environmental Conditions.}
Below, we provide the complete list of extracted environmental conditions extracted from ATR dataset.
\begin{enumerate}
  \item Is there a vehicle present in the image?
  \item Is the terrain mostly flat?
  \item Are there hills or mountains in the background?
  \item Is the sky overcast or cloudy?
  \item Is the image in black and white?
  \item Is there sparse vegetation present in the image?
  \item Does the landscape appear arid or desert-like?
  \item Is there a road or path visible in the image?
  \item Does the image convey a sense of desolation or remoteness?
  \item Is the landscape devoid of human structures?
  \item Is there any evidence of movement, such as tire tracks or dust?
  \item Does the scene have a sense of barrenness or isolation?
  \item Is there a military vehicle like a tank present?
  \item Is there any dust or haze present in the scene?
  \item Is the image devoid of visible human presence?
  \item Is there a single structure visible?
  \item Are there rolling hills or mountains in the background?
  \item Is the landscape described as barren?
  \item Is the lighting subdued or muted?
\end{enumerate}

\subsubsection{Additional Quantitative Analysis.}
\label{app:B_2_3}
\begin{figure}[t]
    \centering
    \includegraphics[width=1\linewidth]{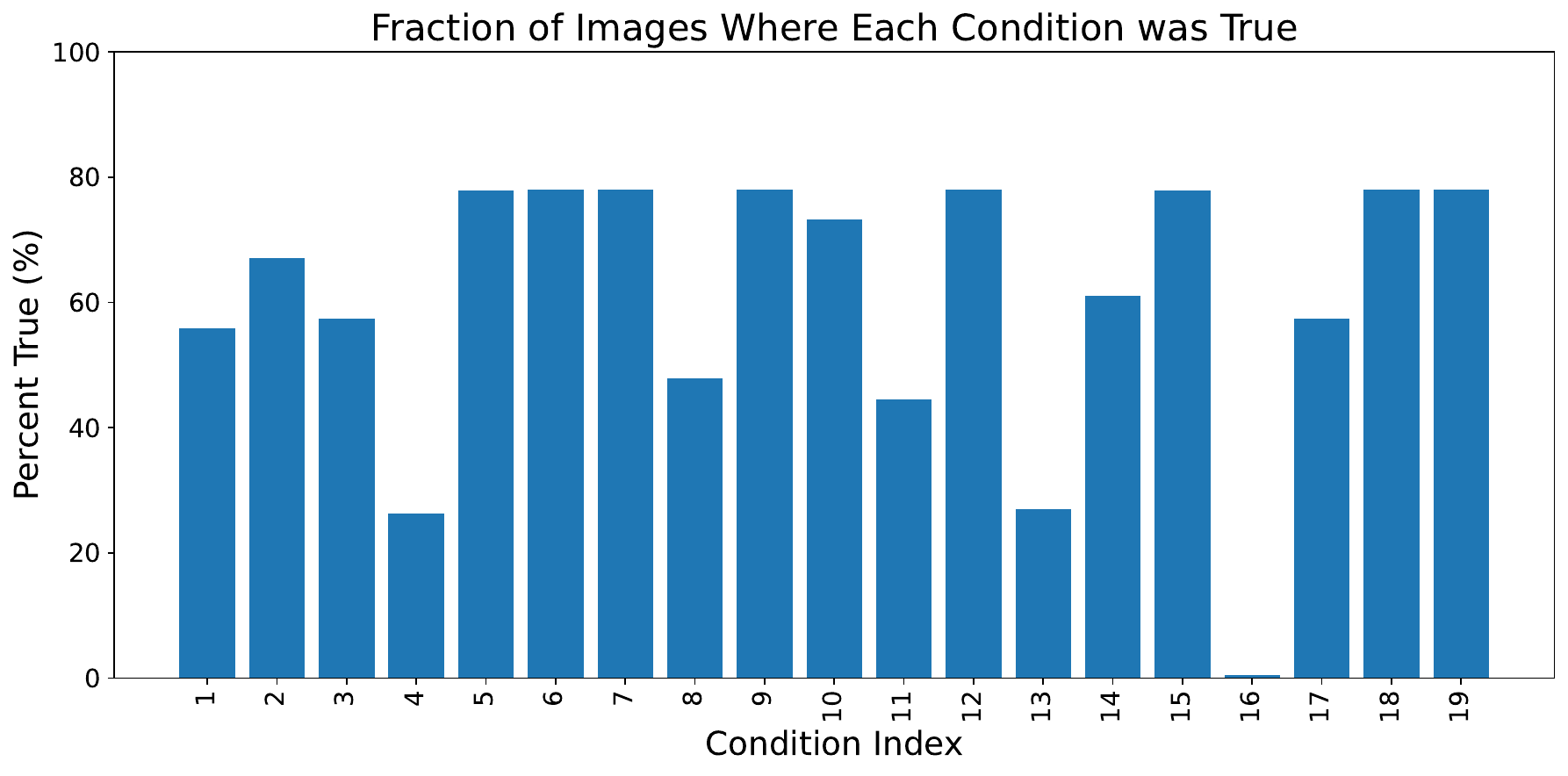}
    \caption{Fraction of images in the seen distances test set for which each condition is true in ATR dataset.}
    \label{fig:atr_seen_fraction_true}
\end{figure}

\begin{figure}[t]
    \centering
    \includegraphics[width=1\linewidth]{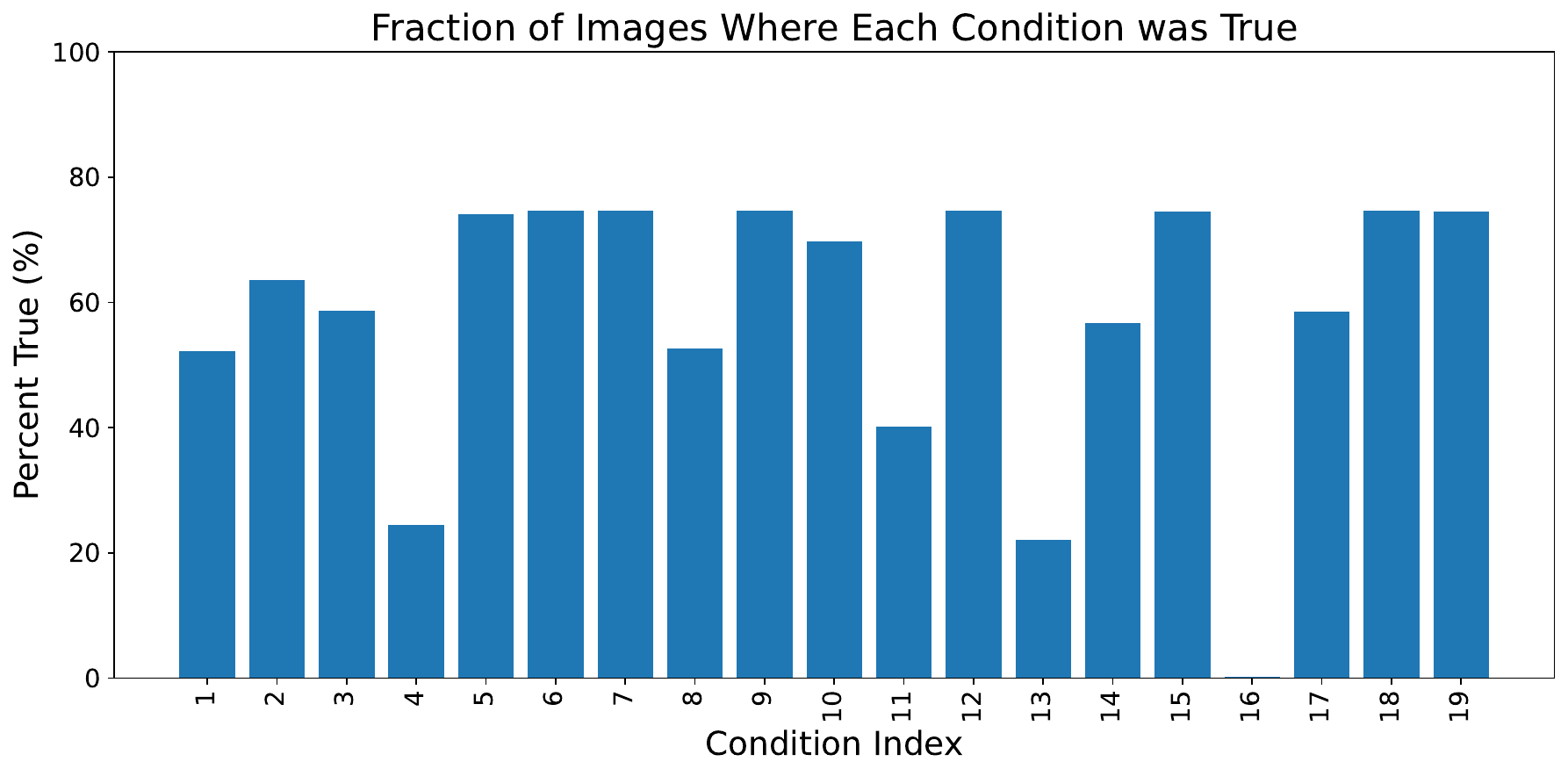}
    \caption{Fraction of images in the unseen distances test set for which each condition is true in ATR dataset.}
    \label{fig:atr_unseen_fraction_true}
\end{figure}

Fig.~\ref{fig:atr_seen_fraction_true} and \ref{fig:atr_unseen_fraction_true} shows the activation of conditions over seen distances and unseen distances test set. We can see that 76\% of the test dataset have at least one active environmental condition.

\section{Prompt Templates}
\label{app:prompts}
In this section, we discuss the prompts used for defining our automatic environmental condition extraction framework. For each step, we use a separate set of system and user prompt defined as:

\textbf{Captioning:}
In this step, we prompt the VLM to describe the images to create image-caption pairs. The prompt template followed is described in Fig.~\ref{fig:captioning_prompt}.

\begin{figure}[t]
    \centering
    \begin{tcolorbox}[colback=gray!10, colframe=gray!50!black, title=System and User prompt template]
    \small
\textbf{System Prompt:} 

``You are an assistant that generates consistent, structured descriptions for the provided image(s). Output should be in the following JSON format:''

\texttt{\{ ``Conditions'': ``<description>''\}}\\
\\
\textbf{User Prompt:} ``Provide a description based on the following image.''\\
\textit{[Image]}

\end{tcolorbox}
    \caption{System and user prompt templates for the VLM “captioning” stage.}
    \label{fig:captioning_prompt}
\end{figure}

\begin{table*}[ht]
    \caption{Class-wise and overall $mAR_{100}$ scores on the ATR dataset (Seen distances). Best and second-best results are highlighted in bold and underline, respectively.}
    \centering
    \resizebox{\linewidth}{!}{\begin{tabular}{l|cccccccccc|c}
    \toprule
    \textbf{Fusion Technique} &
    Pickup & SUV & BTR70 & BRDM2 & BMP2 & T72 & ZSU23 & 2S3 & MTLB & D20 & \textbf{Overall} \\
    \midrule
    Fusion SSD
    & 49.39 & 59.93 & 70.01 & 66.96 & 77.98 & 77.15 & 75.20 & 77.18 & 66.29 & 26.85 & 64.69 \\
    \makecell[l]{Fusion SSD with\\Self-Attention}
    & 46.33 & 55.14 & 64.55 & 62.53 & 72.69 & 70.81 & 71.16 & 72.57 & 58.18 & 0.00 & 57.40 \\
    Learnable Align
    & 56.25 & 59.17 & 70.93 & 65.89 & 78.29 & 77.51 & 76.00 & 77.55 & 65.29 & 21.58 & 64.84 \\
    RGB-X
    & 46.24 & 52.45 & 65.50 & 63.11 & 74.44 & 68.25 & 71.58 & 71.47 & 61.42 & 0.00 & 57.45 \\
    \midrule
    \makecell[l]{VLC Fusion with Human\\ Defined Conditions (n=3)}
    & \textbf{82.06} & \underline{81.53} & \textbf{85.63} & \textbf{83.7} & \textbf{86.98} & \textbf{86.85} & \textbf{85.12} & \textbf{87.75} & \textbf{66.87} & \underline{76.16} & \textbf{82.27} \\
    \makecell[l]{VLC Fusion with Extracted\\ Conditions (n=7)}
    & \underline{80.06} & \textbf{83.26} & \underline{85.06} & \underline{83.34} & \underline{84.15} & \underline{86.46} & \underline{84.83} & \underline{85.68} & \underline{66.49} & \textbf{80.20} & \underline{81.95} \\
    \bottomrule
    \end{tabular}}
    \label{tab:atr_mar_performance_seen}
\end{table*}

\begin{table*}[ht]
    \caption{Class-wise and overall $mAR_{100}$ scores on the ATR dataset (Unseen distances). Best and second-best results are highlighted in bold and underline, respectively.}
    \centering
    \resizebox{\linewidth}{!}{\begin{tabular}{l|cccccccccc|c}
    \toprule
    \textbf{Fusion Technique} &
    Pickup & SUV & BTR70 & BRDM2 & BMP2 & T72 & ZSU23 & 2S3 & MTLB & D20 & \textbf{Overall} \\
    \midrule
    Fusion SSD
    & 1.48 & 9.53 & 18.88 & 19.12 & 21.18 & 22.37 & 15.35 & 30.47 & 8.07 & 2.94 & 14.94 \\
    \makecell[l]{Fusion SSD with\\Self-Attention}
    & 2.25 & 7.61 & 18.33 & 16.56 & 22.74 & 19.99 & 10.73 & 29.27 & 5.14 & 0.36 & 13.30 \\
    Learnable Align
    & 0.08 & 3.02 & 14.71 & 12.13 & 24.93 & 17.41 & 6.25 & 21.15 & 7.07 & 2.84 & 10.96 \\
    RGB-X
    & 0.56 & 8.96 & 16.09 & 12.87 & 18.90 & 22.86 & 13.83 & 28.34 & 6.36 & 0.00 & 12.88 \\
    \midrule
    \makecell[l]{VLC Fusion with Human\\ Defined Conditions (n=3)}
    & \underline{4.09} & \underline{20.88} & \underline{28.76} & \underline{40.64} & \underline{42.39} & \underline{43.31} & \underline{27.22} & \textbf{51.16} & \underline{9.42} & \underline{22.08} & \underline{28.99}\\
    \makecell[l]{VLC Fusion with Extracted\\ Conditions (n=7)}
    & \textbf{16.45} & \textbf{26.36} & \textbf{37.76} & \textbf{46.90} & \textbf{46.58} & \textbf{47.33} & \textbf{44.70} & \underline{44.85} & \textbf{16.43} & \textbf{27.72} & \textbf{35.51} \\
    \bottomrule
    \end{tabular}}
    \label{tab:atr_mar_performance_unseen}
\end{table*}

\textbf{Extraction:}
In this step, we prompt the VLM to provide the set of conditions based on image-caption pairs. The prompt template used is described in Fig.~\ref{fig:extraction_prompt}.

\begin{figure}[t]
    \centering
    \begin{tcolorbox}[colback=gray!10, colframe=gray!50!black, title=System and User prompt template]
    \small
\textbf{System Prompt:} 

``You are an assistant that generates consistent, structured conditions
for the given image. These conditions are based on various aspects
of the image and its description. The conditions should be in the
form of questions. Generate as many unique conditions as possible.
The questions should be in the form of yes/no questions. Do not
include any specific information about the image or description
while generating the conditions. Output should be in the following
JSON format:''

\texttt{\{ "Conditions": [}\\
\texttt{"<condition\_1>",}\\
\texttt{"<condition\_2>"}\\
\texttt{]\}} \\
\\
\textbf{User Prompt:} ``Provide conditions based on the following images and their captions.''\\
\textit{[Images, Captions]}

\end{tcolorbox}
    \caption{System and user prompt templates for the “extraction” stage.}
    \label{fig:extraction_prompt}
\end{figure}

\textbf{Generation:}
In this step, we query the VLM to generate the responses based on the presence and absence of the extracted conditions. The prompt template followed is described in Fig.~\ref{fig:gen_prompt}.

\begin{figure}[t]
    \centering
    \begin{tcolorbox}[colback=gray!10, colframe=gray!50!black, title=System prompt and Input prompt template]
    \small
\textbf{System Prompt:} 

``You are a highly specialized assistant that provides concise answers to specific questions about images, responding to each with either True or False only and returning a JSON object with keys 1 through N corresponding to the question numbers, without any additional context or descriptions.''\\
\\
\textbf{User Prompt:} ``Answer the following questions based on the given image by returning a JSON object with exactly N keys (the strings “1” through “N”), each mapped to a boolean (True or False) corresponding to its question and nothing else; the image is provided after these questions.''\\
\textit{[Question List]}\\

\end{tcolorbox}
    \caption{System and user prompt templates for the “generation” stage.}
    \label{fig:gen_prompt}
\end{figure}

\section{Additional Results from ATR Experiment}
\label{app:atr_results}
In this section, we provide additional results of VLC Fusion and other fusion techniques on ATR dataset. Specifically, we provide the overall and per-class $mAR_{100}$ scores in table~\ref{tab:atr_mar_performance_seen} and \ref{tab:atr_mar_performance_unseen}. As shown, VLC Fusion with extracted conditions performed best in both seen and unseen test scenarios.

\section{Extended Qualitative Examples}
\label{app:qualitative}
\begin{figure*}[t]
    \centering
    \includegraphics[width=1\linewidth]{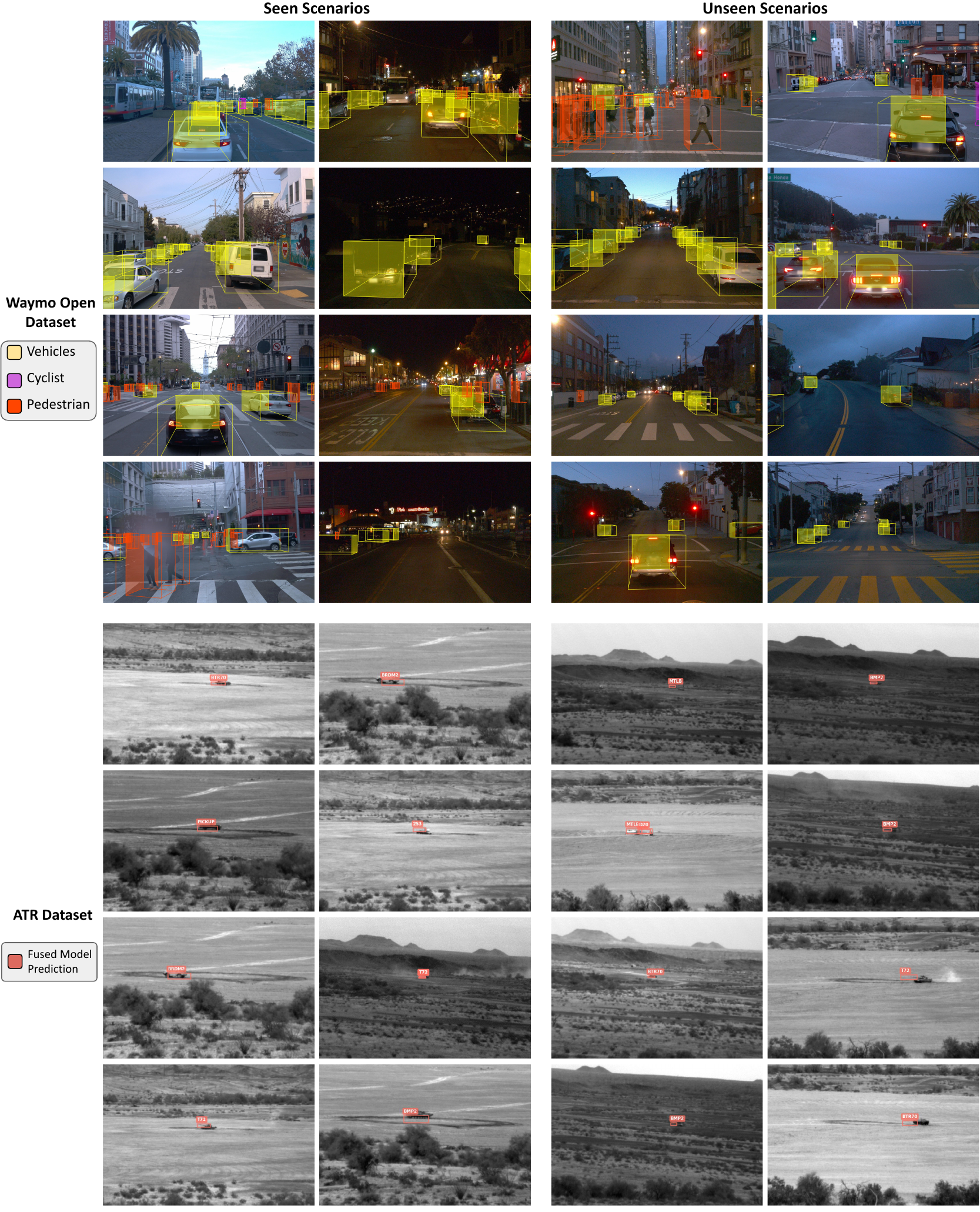}
    \caption{Additional qualitative examples of VLC Fusion for both dataset in seen and unseen scenarios.}
    \label{fig:extended_sample_predictions}
\end{figure*}

In Fig.~\ref{fig:extended_sample_predictions} we provide an extended qualitative examples on object detection performance of VLC Fusion in both dataset, Waymo dataset and ATR dataset, for seen and unseen scenarios.

\end{document}